\documentclass[a4paper]{article}
\usepackage[left=1in,right=1in,top=1in,bottom=1in]{geometry}

\usepackage{amsthm}
\usepackage{amssymb}
\usepackage{amsfonts}
\usepackage{amsmath}
\usepackage{lmodern}
\usepackage[T1]{fontenc}
\usepackage[utf8]{inputenc}

\usepackage{todonotes}
\presetkeys{todonotes}{color=red!30}{}

\usepackage{framed}

\usepackage{graphicx}
\usepackage{caption}
\usepackage{subcaption}
\usepackage{booktabs}

\usepackage{url}

\usepackage{authblk}

\usepackage{tikz}
\usetikzlibrary{shapes.geometric}

\newcommand{\type}[1] {\mbox{\tt \footnotesize {#1}}}

\newcommand{\keys} {{\sf Keys}}
\newcommand{\topics} {{\sf Topics}}

\begin{document}


\title{Domain Modelling in Computational Persuasion\\ for Behaviour Change in Healthcare}


\author[1]{Lisa Chalaguine}
\author[1]{Emmanuel Hadoux}
\author[2]{Fiona Hamilton}
\author[3]{Andrew Hayward}
\author[1]{Anthony Hunter\footnote{Corresponding author (anthony.hunter@ucl.ac.uk)}}
\author[1]{Sylwia Polberg}
\author[4]{Henry W. W. Potts}

\affil[1]{Department of Computer Science, University College London, London, UK}
\affil[2]{eHealth Unit, University College London, London, UK}
\affil[3]{Institute of Epidemiology and Health Care, University College London, London, UK}
\affil[4]{Institute of Health Informatics, University College London, London, UK}

\maketitle


\begin{abstract}
The aim of behaviour change is to help people to change aspects of their behaviour for the better (e.g., to decrease calorie intake, to drink in moderation, to take more exercise, to complete a course of antibiotics once started, etc.). In current persuasion technology for behaviour change, the emphasis is on helping people to explore their issues (e.g., through questionnaires or game playing) or to remember to follow a behaviour change plan (e.g., diaries and email reminders).
However, recent developments in computational persuasion are leading to an argument-centric approach to persuasion that can potentially be harnessed in behaviour change applications. In this paper, we review developments in computational persuasion, and then focus on domain modelling as a key component. We present a multi-dimensional approach to domain modelling. At the core of this proposal is an ontology which provides a representation of key factors, in particular kinds of belief, which we have identified in the behaviour change literature as being important in diverse behaviour change initiatives. Our proposal for domain modelling is intended to facilitate the acquisition and representation of the arguments that can be used in persuasion dialogues, together with meta-level information about them which can be used by the persuader to make strategic choices of argument to present.
\end{abstract}

\begin{quote}
\end{quote}

\begin{quote}
{\bf Declarations of interest:} None
\end{quote}

\section{Introduction}

Persuasion is an activity that involves one party trying to induce another party to believe something or to do something. It is an important and multifaceted human facility. Obviously, sales and marketing is heavily dependent on persuasion. But many other activities involve persuasion such as a doctor persuading a patient to drink less alcohol, a road safety expert persuading drivers to not text while driving, or an online safety expert persuading users of social media sites to not reveal too much personal information online.

The aim of persuasion is for the persuader to change the mind of the persuadee.
Some kinds of interaction surrounding persuasion include:
the persuader collecting information, preferences, etc., from the persuadee;
the persuader providing information, offers, etc., to the persuadee;
and the persuader winning favour (e.g., by flattering the persuadee, by making small talk, by being humorous, etc.).
However, arguments (and counterarguments) are the essential structures for presenting the claims (and counterclaims) in persuasion.
An argument-centric focus on persuasion leads to a number of aspects that can be important in bringing about successful persuasion such as the rationality of the argumentation, appropriateness of the persuader, the appropriateness of the language used in the arguments, the psychological strategies used by the persuader, and the personality of the persuadee.

As computing becomes involved in every sphere of life, so too is persuasion a target for applying computer-based solutions. Persuasion technologies have come out of developments in human-computer interaction research (see, for example, the influential work by Fogg \cite{Fogg98}) with a particular emphasis on addressing the need for systems to help people make positive changes to their behaviour, particularly in healthcare and lifestyle choices. In recent years, a wide variety of systems has been developed to help users to control body weight \cite{LOK10}, to reduce fizzy drink consumption \cite{LOK12}, to increase physical exercise \cite{Z+12}, to decrease stress-related illness \cite{K+10}, to reduce speeding \cite{Maurer16}, and to increase health-conscious food shopping \cite{Siawsolit17}.

Many of these persuasion technologies for behaviour change are based on some combination of questionnaires for finding out information from users, provision of information for directing the users to better behaviour, computer games to enable users to explore different scenarios concerning their behaviour, provision of diaries for getting users to record ongoing behaviour, and messages to remind the user to continue with the better behaviour. These systems tend to be heavily scripted multi-media solutions and they are often packaged as websites and/or apps for mobile devices.

Interestingly, explicit use of argumentation is not central to most current manifestations of persuasion technologies. Either arguments are provided implicitly in the persuasion technology (e.g., through provision of information, or through game playing), or it is assumed that the user has considered the arguments for changing behaviour prior to accessing the persuasion technology (e.g., when using diaries, or receiving email reminders). Explicit argumentation with consideration of arguments and counterarguments is not supported with existing persuasion technologies. Yet, for some tasks in behaviour change, an argument-based approach could be highly beneficial, particularly when someone is lacking some key information, and/or entertaining misconceptions about a topic. Via the use of argument-based technology, personalized context-sensitive information can be provided to fill in those gaps and those misconceptions in a user's information. Overall, the system may be able to change the user's mind about belief in some key arguments, and potentially, as a result, persuade the user to believe and follow up the persuasion goal.

This creates some interesting opportunities for artificial intelligence, using computational models of argument, to develop persuasion technologies for behaviour change where arguments are central. For reviews of computational models of argumentation, a.k.a computational argumentation, see \cite{BD07,BHbook,RS09}. This leads to an opportunity for what we call {\em computational persuasion}, or equivalently, an argument-centric persuasion for behaviour change. Computational models of argument are beginning to offer ways to formalize aspects of persuasion, and with some adaptation and development, they have the potential to be incorporated into valuable tools for behaviour change.

There are numerous potential applications of computational persuasion that could be used to encourage and guide people to change behaviour in healthcare including:

\begin{itemize}

\item {\bf Healthy life-styles} (e.g., eating fewer calories, eating more fruit and veg, doing more exercise, drinking less alcohol).

\item {\bf Treatment compliance} (e.g., undertaking self-management of diabetes, completing a course of antibiotics, completing a course of prophylactics).

\item {\bf Treatment reduction} (e.g., using alternatives to painkillers for premenstrual syndrome, not requesting antibiotics for viral infections)

\item {\bf Problem avoidance} (e.g., taking vaccines, taking malaria prophylactics, using safe sex practices).

\item {\bf Screening} (e.g., participating in breast cancer screening, participating in cervical smear screening, self-screening for prostate cancer, breast cancer, bowel cancer, and melanoma)

\end{itemize}

In an argument-centric approach to persuasion in behaviour change, a software system (the persuader) enters into a dialogue with a user (the persuadee). In the dialogue, arguments and counterarguments can be considered. The system undertakes this dialogue with the aim of providing the right combinations of arguments to persuade the user to accept a specific argument (the persuasion goal) that encapsulates the reason for a change of behaviour in some specific respect. For example, the persuasion goal might be that the user needs to eat fruit in order to be more healthy, and the system presents supporting arguments (based on evidence, expert opinion, explanation of the fit with the user's goals, etc.) and counter-arguments that to correct misconceptions or inconsistencies in the user's goals and opinions.

A key contribution of computational models of argument is a range of formalisms for describing arguments, and relationships between them such as attack (e.g., argument $A$ attacks argument $B$ when $A$ has a claim that negates some of the premises of $B$) and support (e.g., argument $A$ supports argument $B$ when the claim of $A$ implies some of the premises of $B$). For applying an argument-centric approach to persuasion in behaviour change, we also require methods for acquiring and arranging those arguments for a specific domain. Unfortunately, computational models of argument do not provide these kinds of methods, and therefore there is a need to develop methods for domain modelling that will allow us to develop and apply domain models in computational persuasion.

In this paper, our aim is to focus on the role of domain modelling in computational persuasion. We explain how domain modelling is important for an application, we provide a framework for domain modelling based on a multi-dimensional framework for classifying arguments that facilitates the construction of domain models, and we explain how the domain modelling (i.e., the classifications of arguments) can be harnessed to enable good choices of move in a dialogue.

We proceed as follows:
(Section \ref{section:background}) We provide some background to behaviour change theory and identify some key types of factor used in behavioural change models;
(Section \ref{section:computationalpersuasion}) We review computational persuasion for behaviour change;
(Section \ref{section:domainmodelling}) We present a framework for domain modelling that incorporates an ontology based on factors identified in Section \ref{section:background};
(Section \ref{section:casestudies}) We present five cases studies that are based on the framework presented in Section \ref{section:domainmodelling};
(Section \ref{section:use}) We consider how a domain model developed according to the framework presented in Section \ref{section:domainmodelling} can be used by the persuader;
(Section \ref{section:discussion}) We discuss the contributions in the paper, and consider future work.



\section{Behaviour change theory}
\label{section:background}

Behaviour change theory offers a wide range of models that provide connections between the characteristics of a persuadee, and the context of the problem, and the outcomes of the behaviour change initiative (e.g., whether the persuadee is persuaded, whether the persuadee actually changes behaviour, etc.). Different models focus on different aspects of a persuadee and of the context. In general, these models aim to identify characteristics that will have a positive or negative influence on a desired outcome.

These models are often supported by evidence that show for specific behaviour change initiatives, the degree to which the behaviour change initiative was successful. The models can then be applied to new behaviour change initiatives with some indication of the degree to which it might be effective.

\begin{figure}
\centering
 \begin{subfigure}[b]{1\textwidth}
\centering
\begin{tikzpicture}[->,>=latex,thick]
\node (c1) [text centered,text width=4.5cm,shape=rectangle,fill=blue!20,draw] at (0,6) {\scriptsize Severity (e.g., Bowel cancer is a serious illness)};
\node (c2) [text centered,text width=4.5cm,shape=rectangle,fill=blue!20,draw] at (0,4.5) {\scriptsize Susceptibility (e.g., my chances of getting bowel cancer are high)};
\node (c3) [text centered,text width=4.5cm,shape=rectangle,fill=blue!20,draw] at (0,3) {\scriptsize Response effectiveness (e.g., changing my diet would improve my health)};
\node (c4) [text centered,text width=4.5cm,shape=rectangle,fill=blue!20,draw] at (0,1.5) {\scriptsize Self-efficacy (e.g., I am confident that I can change my diet)};
\node (c5) [text centered,text width=4.5cm,shape=rectangle,fill=blue!20,draw] at (0,0) {\scriptsize Fear (e.g., I am scared of getting cancer)};
\node (b) [text centered,text width=2cm,shape=rectangle,fill=red!20,minimum height=3em,draw] at (5,3) {\scriptsize Intentions};
\node (a) [text centered,text width=2cm,shape=rectangle,fill=red!20,minimum height=3em, draw] at (8,3) {\scriptsize Behaviour};
\path	(c1.east) edge[] (b);
\path	(c2.east) edge[] (b);
\path	(c3.east) edge[] (b);
\path	(c4.east) edge[] (b);
\path	(c5.east) edge[] (b);
\path	(b) edge[] (a);
\end{tikzpicture}
\caption{Protection motivation theory.} 
\end{subfigure}
\par\bigskip \par\bigskip
\begin{subfigure}[b]{1\textwidth}
\centering
\begin{tikzpicture}[->,>=latex,thick]
\node (c1) [text centered,text width=4.5cm,shape=rectangle,fill=blue!20,draw] at (0,6) {\scriptsize Belief about outcomes};
\node (c2) [text centered,text width=4.5cm,shape=rectangle,fill=blue!20,draw] at (0,4.5) {\scriptsize Belief about others' attitude to behaviour (e.g., my friends will approve if I lose weight)};
\node (c3) [text centered,text width=4.5cm,shape=rectangle,fill=blue!20,draw] at (0,3) {\scriptsize Motivation to comply with others (e.g., I want the approval of my friends about my healthiness)};
\node (c4) [text centered,text width=4.5cm,shape=rectangle,fill=blue!20,draw] at (0,1.5) {\scriptsize Internal control (skills/ability/information)};
\node (c5) [text centered,text width=4.5cm,shape=rectangle,fill=blue!20,draw] at (0,0) {\scriptsize External control (obstacles/opportunities)};
\node (b1) [text centered,text width=2cm,shape=rectangle,fill=blue!20,minimum height=3em,draw] at (5,5) {\scriptsize Attitude towards the behaviour};
\node (b2) [text centered,text width=2cm,shape=rectangle,fill=blue!20,minimum height=3em,draw] at (5,3) {\scriptsize Subjective norms};
\node (b3) [text centered,text width=2cm,shape=rectangle,fill=blue!20,minimum height=3em,draw] at (5,1) {\scriptsize Behavioural control};
\node (a1) [text centered,text width=2cm,shape=rectangle,fill=red!20,minimum height=3em,draw] at (8,3) {\scriptsize Intentions};
\node (a2) [text centered,text width=2cm,shape=rectangle,fill=red!20,minimum height=3em, draw] at (8,0) {\scriptsize Behaviour};
\path	(c1.east) edge[] (b1);
\path	(c2.east) edge[] (b2);
\path	(c3.east) edge[] (b2);
\path	(c4.east) edge[] (b3);
\path	(c5.east) edge[] (b3);
\path	(b1.east) edge[] (a1);
\path	(b2.east) edge[] (a1);
\path	(b3.east) edge[] (a1);
\path	(b3.east) edge[] (a2.north);
\path	(a1) edge[] (a2);
\end{tikzpicture}
\caption{Theory of planned behaviour.}
\end{subfigure}
\caption{\label{fig:behaviourchange}Two behaviour change models: Protection motivation theory which is based on using health beliefs to predict health behaviours \cite{Rogers75}; And theory of planned behaviour which also considers habits and environmental factors also needed for prediction \cite{Ajzen86}. The examples are from \cite{Ogden11}.}
\end{figure}
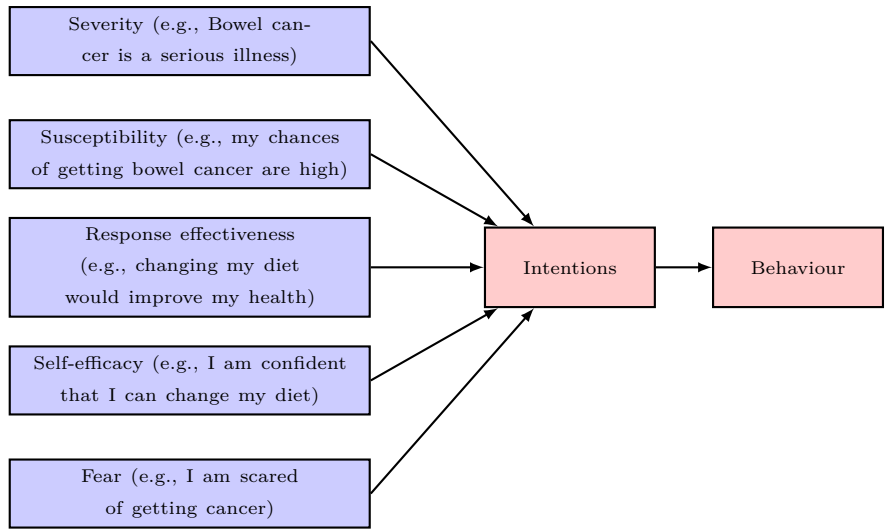
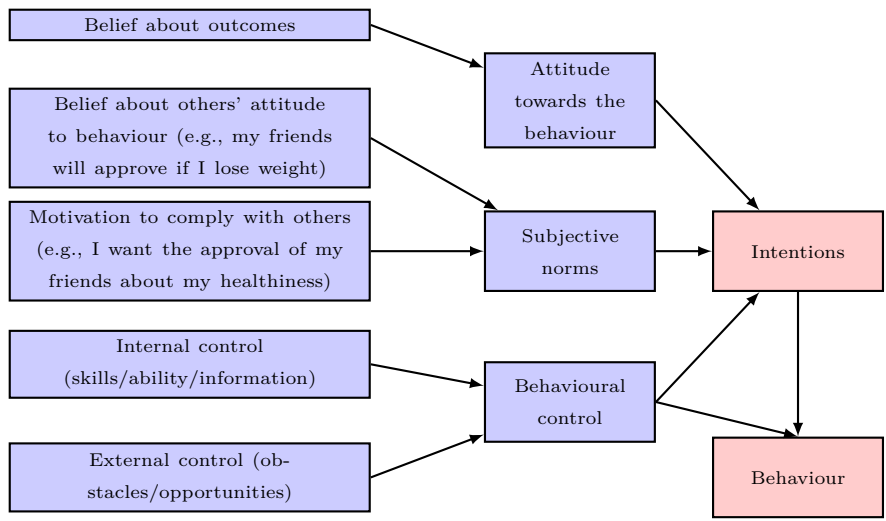

Two important models for behaviour change, which we illustrated in Figure \ref{fig:behaviourchange}, are protection motivation theory \cite{Rogers75} and the theory of planned behaviour \cite{Ajzen86}. For the protection motivation theory, there are five parameters that influence whether or not someone has the intention to change behaviour, and whether or not she does change behaviour. For theory of planned behaviour, there are also some intermediate parameters which influence the intention to change behaviour, and one of which also influences whether or not she does change behaviour. For a review of models of behaviour change, see \cite{Michie14}.

We do not adhere to any specific behaviour change model in this paper. Rather, we would like to identify some interesting dimensions of them that we can harness in our approach to domain modelling. We start with how the system (persuader) or user (persuadee) would like the world to be.

\begin{itemize}

\item {\bf Goals of the agent that are related to the healthcare problem}: Whether an agent has short or long term goals that impact on the problem or are impacted by the problem (see, for instance, the {\em goals} parameter of self-determination theory \cite{Deci2012}). For example, an agent may have a short-term goal to get fit in order to run a marathon, or a long-term goal to remain healthy in order to see her grandchildren grow up. Note, the goals of a persuader are not necessarily the same as the persuadee. For example, a persuader might have the goal for the persuadee to eat more fruit and less chocolate, and the persuadee might have the goal to indulge in comfort foods such as chocolate.

\end{itemize}

Now we consider how the world is perceived by either the system or the user. For an agent that could be subject to a behaviour change initiative, the beliefs of the agent and about the agent are regarded as of central importance in a number of behaviour change models. By beliefs we mean not only about beliefs about facts but also beliefs about capability, opportunity, and motivation \cite{Michie14}. There is substantial evidence in the behaviour change literature that shows the importance of the beliefs of a persuadee affecting the likelihood that the persuadee will be persuaded by a specific behaviour change intervention. We draw out some of the kinds of belief that are important in the following listing where we consider the beliefs with respect to some healthcare problem. Note, these kinds of beliefs are not meant to be exhaustive nor disjoint.


\begin{itemize}

\item {\bf Beliefs concerning causes of the healthcare problem}: Whether the agent regards the cause being internal (i.e., herself) or external (see, for instance, the {\em perceived control} parameter of the health belief model \cite{Rosenstock1988} or the modelling of locus of control \cite{Wallston1982}). For example, an agent acknowledges that she is overweight because she drinks too many soft drinks, versus the agent who blames the food corporations for making the drinks too calorific.

\item {\bf Beliefs concerning opportunities for resolving the healthcare problem}: Whether the agent is aware of options for addressing the problem (see, for instance, the {\em external control} parameter in the theory of planned behaviour \cite{Ajzen86}). For example, for an agent who has an alcohol consumption problem, an option for attempting to resolve the problem is a self-help group for encouraging drinking in moderation.

\item {\bf Beliefs concerning obstacles for resolving the healthcare problem}: Whether the agent perceives reasons for being unable to address the problem (see, for instance, the {\em external control} parameter in the theory of planned behaviour \cite{Ajzen86}). For example, for an agent who has an alcohol consumption problem, an obstacle for attempting to resolve the problem might be that the agent works in a bar.

\item {\bf Beliefs concerning capacity for resolving the healthcare problem}: Whether the agent perceives himself as having the capability for resolving the problem (see, for instance, the {\em self-efficacy} parameter in protection motivation theory \cite{Rogers75} or the {\em internal control parameter} in the theory of planned behaviour \cite{Ajzen86}). For example, for an agent who has an alcohol consumption problem, and who is aware of a self-help group for encouraging drinking in moderation, believes she has the motivation to go to the self-help group regularly, and to follow through with a serious attempt at drinking in moderation. As another example, an agent might believe that she is unable to follow a specific diet aimed at reducing calorie intake (i.e., she believes the temptation to deviate from the diet would be too great). Beliefs concerning the past behaviour of the agent may be an important part of this: Whether an agent undertakes actions to compound the problem, and/or whether the agent undertakes actions that resolve the problem. For example, for an agent who has decided to attend a self-help group for encouraging drinking in moderation, what is her current rate of alcohol consumption, and how often does she attend the self-help group? Past behaviour has been used to augment the theory of planned behaviour model to predict young people's sun safety in Australia \cite{White2008}.

\item {\bf Beliefs concerning risks from the healthcare problem}: Whether an agent has a reasonable or biased perception of her own risk concerning the healthcare problem (see, for instance, the {\em susceptibility} parameter in protection motivation theory \cite{Rogers75} or the {\em susceptibility} or {\em severity} parameters of the health belief model \cite{Rosenstock1988} or the {\em threat} parameter of the health action process approach \cite{Schwarzer2008}). For example, an agent might think that drinking a large quantity of alcohol at the weekend is ok if she does not drink during the week. For another example, an agent might think she is not at risk from cervical cancer because she is not sexually active.

\item {\bf Beliefs concerning benefits from resolving the healthcare problem}: Whether an agent has a reasonable or biased perception of her own benefits concerning the healthcare problem (see, for instance, the {\em belief about outcomes} parameter in the theory of planned behaviour \cite{Ajzen86} or the {\em susceptibility} or {\em severity} parameters of the health belief model \cite{Rosenstock1988}). For example, a smoker might be unaware that there are short-term health benefits (decreased risk of cardiac arrest) and long term health benefits (decreased risk of cancer) of ceasing to smoke even if she has been a smoker for many years.

\item {\bf Beliefs concerning costs of resolving the healthcare problem}: Whether an agent has a reasonable or biased perception of the cost of undertaking a course of action to resolve the healthcare problem (see, for instance, the {\em cost} parameter of the health belief model \cite{Rosenstock1988}). For example, an unemployed person might believe that she cannot participate in exercise activities at her local sports centre because she lacks money, and so she might be unaware that often sports centres offer reduced cost or free activities for unemployed people. For another example, an agent might believe that it is cool to smoke, and hence if she stops smoking, this will be a price to pay (i.e., a cost). \

\item {\bf Beliefs concerning the motivation of the agent for resolving the healthcare problem}: Whether an agent has a high or low motivation for addressing a healthcare problem (see, for instance, the {\em motivation} parameters of the health belief model \cite{Rosenstock1988} or the {\em motivation} parameter of the information-motivation-behavioural skills model \cite{Fisher2003}). For example, is an agent just considering joining a gym because the idea is vaguely appealing, or is she seriously interested in finding a course of action to improve her fitness? Motivation captures the drive or effort an agent may put into pursuing a goal, or a addressing a specific commitment. For example, a pair of friends may make the commitment to join a local fun run. The first agent may be very unfit but goes for a training run 5 days a week in preparation for the fun run, whereas the second agent may be fitter than the first agent but does no training. In this case, the first agent is showing more commitment than the second agent.

\item {\bf Beliefs concerning the agent's community and their view of the healthcare problem}: Whether an agent believes that there are others in her community who have views on the agent and/or on the healthcare problem (see, for instance, the {\em belief's about others' attitude to behaviour} or the {\em motivation to comply with others} parameters in the theory of planned behaviour \cite{Ajzen86} or the {\em social influences} parameter of the integrated change model \cite{DeVries2005}). For example, an agent might believe that her family thinks she drinks too much, and that they would be happy if she reduced her consumption to a more healthy level. As another example, an agent might have the desire to lose weight because she wants to be seen as healthy by her friends.

\item {\bf Beliefs concerning commitments of the agent that are related to the healthcare problem}: Whether an agent has really committed to address a specific healthcare problem and to a specific course of action for addressing that problem. A commitment is an action that goes some way to addressing a specific goal \cite{Lokhorst2013}. For example, an agent may have a goal to get fit, but the agent then needs to commit to getting fit, and to a specific regime for getting fit such as going for a 20 minute jog three times a week. As another example, an agent may have a goal of running a marathon, and part of addressing that goal, she may have paid the fee to enter the London Marathon (i.e., paying the fee is a commitment). A commitment can also support the case for new goals. For example, if an agent is not training regularly, and she pays the fee to enter a marathon, then this commitment supports the goal that she should start to train regularly.



\end{itemize}

The options we discuss above are some key dimensions that we can harness in computational persuasion, and in particular, we use them to help us construct domain models. As we will explain in the following sections, they provide cues for arguments and counterarguments that we can include in domain models, and they provide meta-level typing of those arguments and counterarguments.



\section{Towards computational persuasion in behaviour change}
 \label{section:computationalpersuasion}

We now turn to how we can bring an argument-based approach into software for persuasion, thereby leading to a notion of computational persuasion. We start by defining the notion of an automated persuasion system, and use this to define the notion of computational persuasion. We then consider our requirements, in the short-term, and components, for an automated persuasion system in healthcare.

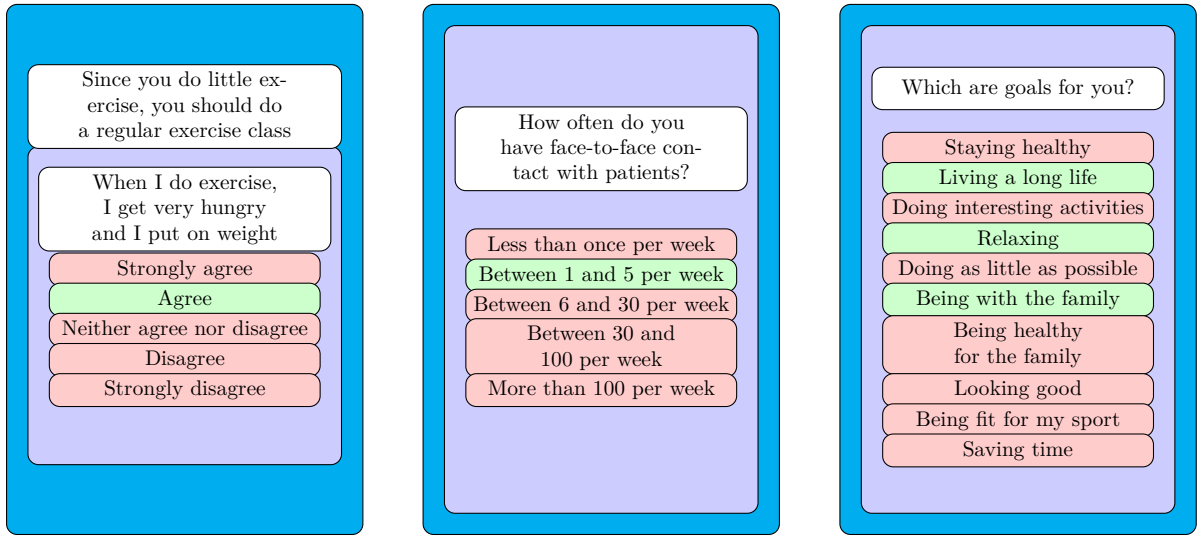
\begin{figure}
 \centering
 \begin{subfigure}[b]{0.3\textwidth}
\centering
\begin{tikzpicture}[scale=0.8,every node/.style={scale=0.8}]
\node[rectangle,draw,fill=cyan,text width=16em, text centered, rounded corners,minimum height=25em] (A) at (3,4) {};
\node[rectangle,draw,fill=blue!20,text width=14em, text centered, rounded corners,minimum height=15em] (A) at (3,3.4) {};
\node[rectangle,draw,fill=white,text width=14em, text centered, rounded corners,minimum height=2em] (A) at (3,6.7) {Since you do little exercise, you should do a regular exercise class};
\node[rectangle,draw,fill=white,text width=13em, text centered, rounded corners,minimum height=2em] (A) at (3,5) {When I do exercise, I get very hungry and I put on weight};
\node[rectangle,draw,fill=red!20,text width=12em, text centered, rounded corners,minimum height=0.8em] (A) at (3,4) {Strongly agree};
\node[rectangle,draw,fill=green!20,text width=12em, text centered, rounded corners,minimum height=0.6em] (A) at (3,3.5) {Agree};
\node[rectangle,draw,fill=red!20,text width=12em, text centered, rounded corners,minimum height=1em] (A) at (3,3) {Neither agree nor disagree};
\node[rectangle,draw,fill=red!20,text width=12em, text centered, rounded corners,minimum height=1em] (A) at (3,2.5) {Disagree};
\node[rectangle,draw,fill=red!20,text width=12em, text centered, rounded corners,minimum height=1em] (A) at (3,2) {Strongly  disagree};
\end{tikzpicture}
  \caption{Asking for argument belief.}
  \label{fig:interface1}
 \end{subfigure}
 \quad
 ~ 
 \begin{subfigure}[b]{0.3\textwidth}
\centering
\begin{tikzpicture}[scale=0.8,every node/.style={scale=0.8}]
\node[rectangle,draw,fill=cyan,text width=16em, text centered, rounded corners,minimum height=25em] (A) at (3,4) {};
\node[rectangle,draw,fill=blue!20,text width=14em, text centered, rounded corners,minimum height=23em] (A) at (3,4) {};
\node[rectangle,draw,fill=white,text width=13em, text centered, rounded corners,minimum height=2em] (A) at (3,6) {How often do you have face-to-face contact with patients?};
\node[rectangle,draw,fill=red!20,text width=12em, text centered, rounded corners,minimum height=0.8em] (A) at (3,4.4) {Less than once per week};
\node[rectangle,draw,fill=green!20,text width=12em, text centered, rounded corners,minimum height=0.6em] (A) at (3,3.9) {Between 1 and 5 per week};
\node[rectangle,draw,fill=red!20,text width=12em, text centered, rounded corners,minimum height=1em] (A) at (3,3.4) {Between 6 and 30 per week};
\node[rectangle,draw,fill=red!20,text width=12em, text centered, rounded corners,minimum height=1em] (A) at (3,2.7) {Between 30 and 100 per week};
\node[rectangle,draw,fill=red!20,text width=12em, text centered, rounded corners,minimum height=1em] (A) at (3,2) {More than 100 per week};
\end{tikzpicture}
  \caption{Asking for a fact.}
  \label{fig:interface2}
 \end{subfigure}
 \quad
 ~ 
 \begin{subfigure}[b]{0.3\textwidth}
\centering
\begin{tikzpicture}[scale=0.8,every node/.style={scale=0.8}]
\node[rectangle,draw,fill=cyan,text width=16em, text centered, rounded corners,minimum height=25em] (A) at (3,4) {};
\node[rectangle,draw,fill=blue!20,text width=14em, text centered, rounded corners,minimum height=23em] (A) at (3,4) {};
\node[rectangle,draw,fill=white,text width=13em, text centered, rounded corners,minimum height=2em] (A) at (3,7) {Which are goals for you?};
\node[rectangle,draw,fill=red!20,text width=12em, text centered, rounded corners,minimum height=0.8em] (A) at (3,6) {Staying healthy};
\node[rectangle,draw,fill=green!20,text width=12em, text centered, rounded corners,minimum height=0.6em] (A) at (3,5.5) {Living a long life};
\node[rectangle,draw,fill=red!20,text width=12em, text centered, rounded corners,minimum height=1em] (A) at (3,5) {Doing interesting activities};
\node[rectangle,draw,fill=green!20,text width=12em, text centered, rounded corners,minimum height=1em] (A) at (3,4.5) {Relaxing};
\node[rectangle,draw,fill=red!20,text width=12em, text centered, rounded corners,minimum height=1em] (A) at (3,4) {Doing as little as possible};
\node[rectangle,draw,fill=green!20,text width=12em, text centered, rounded corners,minimum height=1em] (A) at (3,3.5) {Being with the family};
\node[rectangle,draw,fill=red!20,text width=12em, text centered, rounded corners,minimum height=1em] (A) at (3,2.75) {Being healthy for the family};
\node[rectangle,draw,fill=red!20,text width=12em, text centered, rounded corners,minimum height=1em] (A) at (3,2) {Looking good};
\node[rectangle,draw,fill=red!20,text width=12em, text centered, rounded corners,minimum height=1em] (A) at (3,1.5) {Being fit for my sport};
\node[rectangle,draw,fill=red!20,text width=12em, text centered, rounded corners,minimum height=1em] (A) at (3,1) {Saving time};
\end{tikzpicture}
  \caption{Asking for goals.}
  \label{fig:interface3}
 \end{subfigure}
 \caption{Interface for an asymmetric dialogue move for asking the user's belief in an argument. (a) The top argument is by the APS, and the second argument is a counterargument presented by the APS. The user uses the menu to give her belief in the counterargument. (b) A query is asked that may be used in a user model and menu of answers is provided. (c) A query is asked by the system to determine the goals of the user. Here the user may select number of the items on the list.}\label{fig:interface}
\end{figure}

An {\bf automated persuasion system} (APS) is a system that can engage in a dialogue with a user (the persuadee) in order to persuade the persuadee to do (or not do) some action or to believe (or not believe) something \cite{Hunter2016comma}. To do this, an APS aims to use convincing arguments in order to persuade the persuadee. The dialogue may involve moves including queries, claims, and importantly, arguments and counterarguments, that are presented according to some protocol that specifies which moves are permitted or obligatory at each dialogue step. The dialogue may involve stages where the system finds out more about the persuadee's beliefs, intentions and desires, and where the system offers arguments with the aim of changing the persuadee's beliefs, intentions and desires. The system also needs to handle objections or doubts (represented by counterarguments) with the aim of convincing the user to accept the persuasion goal (i.e., the argument that encapsulates the reason for a change of behaviour in some respect).

The dialogue may be asymmetric since the kinds of moves that the APS can present may be different to the moves that the persuadee may make. For instance, the persuadee might be restricted to only making arguments by selecting them from a menu (in order to obviate the need for natural language processing of arguments being entered). In the extreme, it may be that only the APS can make moves.
In Figure \ref{fig:interface1}, a dialogue step is illustrated where a user can state the degree of agreement or disagreement in an argument, and Figure \ref{fig:interface2} and Figure \ref{fig:interface3} illustrate dialogue moves that involve queries.

Whether an argument is convincing depends on the context, and on the characteristics of the persuadee. An APS maintains a model of the persuadee to predict what arguments and counterarguments the persuadee knows about and/or believes, and this can be harnessed by the strategy of the APS in order to choose good moves to make in the dialogue.

Since, we assume that an APS for behaviour change is a software application running on a desktop or mobile device,
this creates some difficult short-term challenges to automate persuasion. We summarize these challenges as follows.

\begin{enumerate}
\item Need for informative asymmetric dialogues without natural language interface.
\item Need for short dialogues to keep the user engaged.
\item Need for well-chosen arguments to maximize impact of the user.
\item Need to model the user beliefs in order to be able to optimize the dialogue.
\item Need to learn from previous interactions with the agent or similar agents.
\item Need to model the domain to generate arguments/counterarguments to present to the user.
\end{enumerate}

In the rest of this section, we discuss how the requirements can be addressed. Some of our proposals are preliminary as these are challenging research questions.

{\bf Computational persuasion} is the study of formal models of dialogues involving arguments and counterarguments, user models, and strategies, for APSs \cite{Hunter2016comma}. As such, computational persuasion builds on developments in computational models of argument which is a branch of AI concerned with formalizing aspects of the human ability to construct, exchange, and evaluate argumentation and counterarguments.





To illustrate how a dialogue can lead to the presentation of an appropriate context-sensitive argument consider the example in Table \ref{tab:exampledialogue}. In this, only the APS presents arguments, and when it is the user's turn she can only answer questions (e.g., yes/no questions) or select arguments from a menu.

\begin{table}
\centering
\begin{tabular}{lll}
\toprule
Step & Who & Move \\
\midrule
1 & APS & To improve your health, you could join an exercise class\\
2 & User & Exercise classes are boring\\
3 & APS & There are exciting exercise classes. You can do an indoor climbing course.\\
4 & User & It is too expensive\\
5 & APS & Do you work?\\
6 & User & No\\
7 & APS & If you are registered unemployed, then the local sports centre offers \\
&& a free indoor climbing course\\
8 & APS & Would you try this?\\
9 & User & Yes\\
\bottomrule
\end{tabular}
\caption{\label{tab:exampledialogue}Simple example of an asymmetric dialogue between a user and an APS. As no natural language processing is assumed, the arguments posted by the user are actually selected by the user from a menu provided by the APS.}
\end{table}

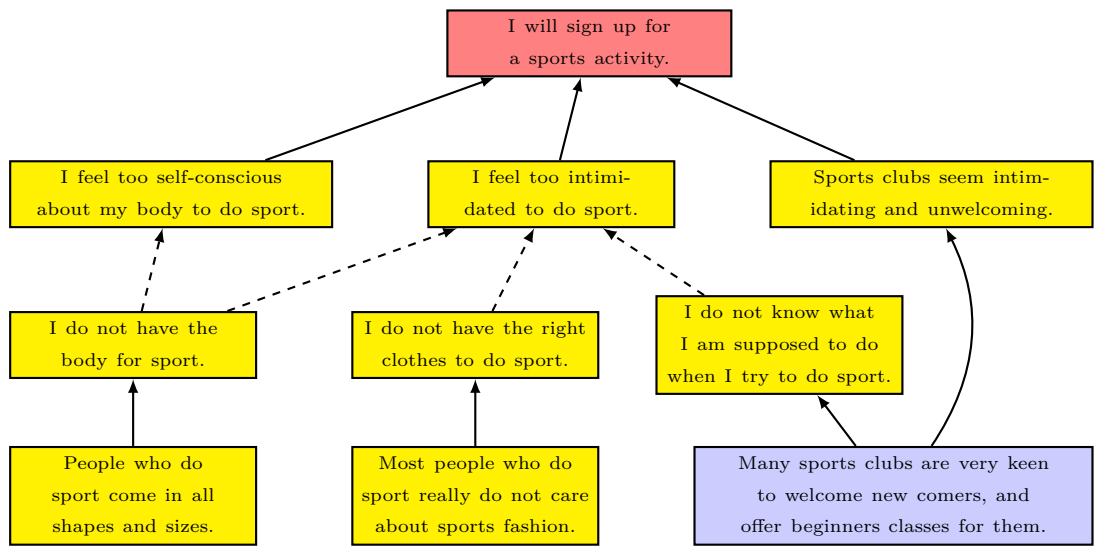
\begin{figure}
  \centering
\begin{tikzpicture}[->,>=latex,thick]
\node (pg) [text centered,text width=3.5cm,shape=rectangle,fill=red!50,draw] at (6,8) {\scriptsize I will sign up for a sports activity.};
\node (a1) [text centered,text width=4cm,shape=rectangle,fill=yellow,draw] at (0.5,6) {\scriptsize I feel too self-conscious about my body to do sport.};
\node (a2) [text centered,text width=3cm,shape=rectangle,fill=yellow,draw] at (5.5,6) {\scriptsize I feel too intimidated to do sport.};
\node (a3) [text centered,text width=4cm,shape=rectangle,fill=yellow,draw] at (10.5,6) {\scriptsize Sports clubs seem intimidating and unwelcoming.};
\path[]	(a1) edge[] node[] {} (pg);
\path[]	(a2) edge[] node[] {} (pg);
\path[]	(a3) edge[] node[] {} (pg);
\node (b1a) [text centered,text width=3cm,shape=rectangle,fill=yellow,draw] at (0,4) {\scriptsize I do not have the body for sport.};
\path[]	(b1a) edge[dashed] node[] {} (a1);
\path[]	(b1a) edge[dashed] node[] {} (a2);
\node (b2a) [text centered,text width=3cm,shape=rectangle,fill=yellow,draw] at (4.5,4) {\scriptsize I do not have the right clothes to do sport.};
\node (b2b) [text centered,text width=3cm,shape=rectangle,fill=yellow,draw] at (8.5,4) {\scriptsize I do not know what I am supposed to do when I try to do sport.};
\path[]	(b2a) edge[dashed] node[] {} (a2);
\path[]	(b2b) edge[dashed] node[] {} (a2);
\node (c1a) [text centered,text width=3cm,shape=rectangle,fill=yellow,draw] at (0,2) {\scriptsize People who do sport come in all shapes and sizes.};
\path[]	(c1a) edge[] node[] {} (b1a);
\node (c2a) [text centered,text width=3cm,shape=rectangle,fill=yellow,draw] at (4.5,2) {\scriptsize Most people who do sport really do not care about sports fashion.};
\path[]	(c2a) edge[] node[] {} (b2a);
\node (c3a) [text centered,text width=5cm,shape=rectangle,fill=blue!20,draw] at (10,2) {\scriptsize Many sports clubs are very keen to welcome new comers, and offer beginners classes for them.};
\path[]	(c3a) edge[] node[] {} (b2b);
\path[]	(c3a) edge[bend right] node[] {} (a3);
\end{tikzpicture}
\caption{\label{fig:topic}An example of a bipolar argument graph (i.e. a graph where each node denotes an argument and each arc denotes either a support or attack relationship) concerning persuasion to sign up for a sports activity. A dashed line denotes a support (where the source of the arc is a supporter and the target of the arc is the supportee) and a solid line denotes an attack  (where the source of the arc is an attacker and the target of the arc is the attackee) .}
\end{figure}

In order to provide this behaviour, we assume an APS has the following components that are used to operate in a persuasion dialogue.

\begin{description}

\item[Domain model] This contains the arguments that can be presented in the dialogue by the system, and it also contains the arguments that the user may entertain. The domain model can be represented by a bipolar argument graph (see, for example, Figure \ref{fig:topic}). This is a graph where each node is an argument, and each arc denotes a relationship between pairs of arguments. We consider two types of relationship for an arc from A to B. The first is an attack relationship, and so the arc from A to B denotes that A attacks B (i.e., A is a counterargument for B). The second is a support relationship, and so the arc from A to B denotes that A supports B (i.e., A provides further information that supports for B).

\item[User model] This contains information about the user that can be used by the system for making good choices of move. The information in the user model is what the system believes is true about the user.
A key dimension that we consider in the user model is the belief that the user may have in the arguments, and as the dialogue proceeds, the model can be updated \cite{Hunter2015ijcai} based on the results of the queries and of the arguments posited. Other dimensions of a user model may include some categorization of the personality of the user, the topics the user may be interested, the viewpoints the user may hold, the goals of the user, etc.

\item[Protocol] A dialogue is a sequence of moves such as asking a query, making a claim, presenting a premises, conceding to a premise presented by another agent, etc. The protocol specifies the moves that are allowed or required by each participant at step of a dialogue. There are a number of proposals for dialogues (e.g., \cite{Pra05,Pra06,FT11,CP12}).
For examples of protocols for persuasion in behaviour change, see \cite{Hunter2015ijcai,Hunter2016sum}. The dialogue may involve steps where the system finds out more about the user's beliefs, intentions and desires, and where the system offers arguments with the aim of changing the user's beliefs, intentions and desires. Moves can involve arguments taken from the domain model, and/or they can be queries to improve the domain model.

\item[Strategy model] The strategy model uses the user model to select the moves that are allowed by the protocol. There are a number of roles for arguments. For instance, an argument may be a persuasion goal (i.e., an argument that the system wants the user to accept), or a user counterargument (i.e., an argument that the user regards as a counterargument against an argument by the system), or a system counterargument (i.e., an argument that the system regards as a counterargument against an argument by held by the system), or a user supporting argument (i.e., an argument that the user regards as supporting an argument by held the user), or a system supporting argument (i.e., an argument that the system regards as supporting an argument presented by the system). We can harness decision-theoretic decision rules for optimizing the choice of arguments based on the user model \cite{HadouxHunter2017}.
See \cite{Thimm14} for a review of strategies in multi-agent argumentation.

\end{description}





There are some promising developments that could form the basis of APSs for behaviour change. Furthermore, there have already been some promising studies using dialogue games
for health promotion \cite{Grasso1998,Cawsey1999,Grasso2000,Grasso2003},
embodied conversational agents for encouraging exercise \cite{Nguyen2007},
dialogue management for persuasion \cite{Andrews2008},
and tailored assistive living systems for encouraging exercise \cite{Guerrero2016},
that indicate the potential for APSs.

Since our primary concern in the paper is in domain modelling, our focus will be on how the domain model can be structured (Sections \ref{section:domainmodelling} and \ref{section:casestudies}). However, we will also touch on how this structure can influence the design of the user model and how this structure can be harnessed with the user model for making strategic choices of move in a dialogue (Section \ref{section:use}).

\section{Framework for domain modelling}
 \label{section:domainmodelling}

The aim of domain modelling for an APS is to catalogue a range of arguments that can be deployed in a persuasion dialogue, and to annotate each of those arguments with meta-level information about them. This meta-level information can help to ensure that the structure of graph is adequate (e.g., arguments that are based on incorrect, but commonly held beliefs, are attacked by factual arguments), and more importantly they can be used to make strategic choices of move. For instance, the most appropriate arguments can be selected according to what is known about the persuadee.

\subsection{Ontological types}
\label{section:ontological}

We start by introducing a number of ontological types. These types provide a breakdown of the kinds of belief introduced in our review of factors in behaviour change models in Section \ref{section:background}. Our approach is to have an ontology that identifies whether the ontological type (and therefore an argument that has this type) concerns the healthcare problem or the healthcare solution or both.
Note, the types are not intended to be exhaustive or disjoint. This means that some arguments may be tagged with multiple ontological types.

\begin{figure}
\centering
\begin{tikzpicture}[->]

\node (cause) [text centered,text width=25mm,shape=rectangle,fill=yellow,draw] at (0,10) {\footnotesize {\bf Cause} of the problem};

\node (risk) [text centered,text width=25mm,shape=rectangle,fill=yellow,draw] at (0,8) {\footnotesize {\bf Risk} from the problem};

\node (benefit) [text centered,text width=25mm,shape=rectangle,fill=yellow,draw] at (0,6) {\footnotesize {\bf Benefit} from resolving the problem};

\node (motivation) [text centered,text width=25mm,shape=rectangle,fill=yellow,draw] at (0,4) {\footnotesize {\bf Motivation} for resolving the problem};

\node (commitment) [text centered,text width=25mm,shape=rectangle,fill=yellow,draw] at (0,2) {\footnotesize {\bf Commitment} for resolving the problem};


\node (problem) [text centered,text width=2cm,shape=rectangle,fill=red!50,draw] at (4,6) {\footnotesize {\bf PROBLEM}};


\node (community) [text centered,text width=25mm,shape=rectangle,fill=yellow,draw] at (6,10) {\footnotesize {\bf Community} view of the problem and its resolution};

\node (attitude) [text centered,text width=25mm,shape=rectangle,fill=yellow,draw] at (6,8) {\footnotesize {\bf Attitude} to considering the problem and/or solution};

\node (myth) [text centered,text width=25mm,shape=rectangle,fill=yellow,draw] at (6,4) {\footnotesize {\bf Myth} concerning problem and/or solution};

\node (background) [text centered,text width=25mm,shape=rectangle,fill=yellow,draw] at (6,2) {\footnotesize {\bf Background} to problem and/or solution};


\node (solution) [text centered,text width=2cm,shape=rectangle,fill=green!50,draw] at (8,6) {\footnotesize {\bf SOLUTION}};


\node (opportunity) [text centered,text width=25mm,shape=rectangle,fill=yellow,draw] at (12,10) {\footnotesize {\bf Opportunity} for a solution to the problem};

\node (cost) [text centered,text width=25mm,shape=rectangle,fill=yellow,draw] at (12,8) {\footnotesize {\bf Cost} of solution};

\node (obstacle) [text centered,text width=25mm,shape=rectangle,fill=yellow,draw] at (12,6) {\footnotesize {\bf Obstacle} to applying the solution};

\node (sideeffect) [text centered,text width=25mm,shape=rectangle,fill=yellow,draw] at (12,4) {\footnotesize {\bf Side-effect} from applying the solution};

\node (capacity) [text centered,text width=25mm,shape=rectangle,fill=yellow,draw] at (12,2) {\footnotesize {\bf Capacity} for applying the solution};


\path[]	(cause.east) edge[] node[] {} (problem.west);
\path[]	(risk.east) edge[] node[] {} (problem.west);
\path[]	(benefit.east) edge[] node[] {} (problem.west);
\path[]	(motivation.east) edge[] node[] {} (problem.west);
\path[]	(commitment.east) edge[] node[] {} (problem.west);

\path[]	(opportunity.west) edge[] node[] {} (solution.east);
\path[]	(cost.west) edge[] node[] {} (solution.east);
\path[]	(obstacle.west) edge[] node[] {} (solution.east);
\path[]	(sideeffect.west) edge[] node[] {} (solution.east);
\path[]	(capacity.west) edge[] node[] {} (solution.east);

\path[]	(community.west) edge[bend right] node[] {} (problem.north);
\path[]	(community.east) edge[bend left] node[] {} (solution.north);

\path[]	(attitude.west) edge[bend right] node[] {} (problem.north);
\path[]	(attitude.east) edge[bend left] node[] {} (solution.north);

\path[]	(myth.west) edge[bend left] node[] {} (problem.south);
\path[]	(myth.east) edge[bend right] node[] {} (solution.south);

\path[]	(background.west) edge[bend left] node[] {} (problem.south);
\path[]	(background.east) edge[bend right] node[] {} (solution.south);

\end{tikzpicture}
\caption{\label{fig:ontology}Graphical representation of the ontological types and their relationship to either the healthcare problem or healthcare solution.}
\end{figure}
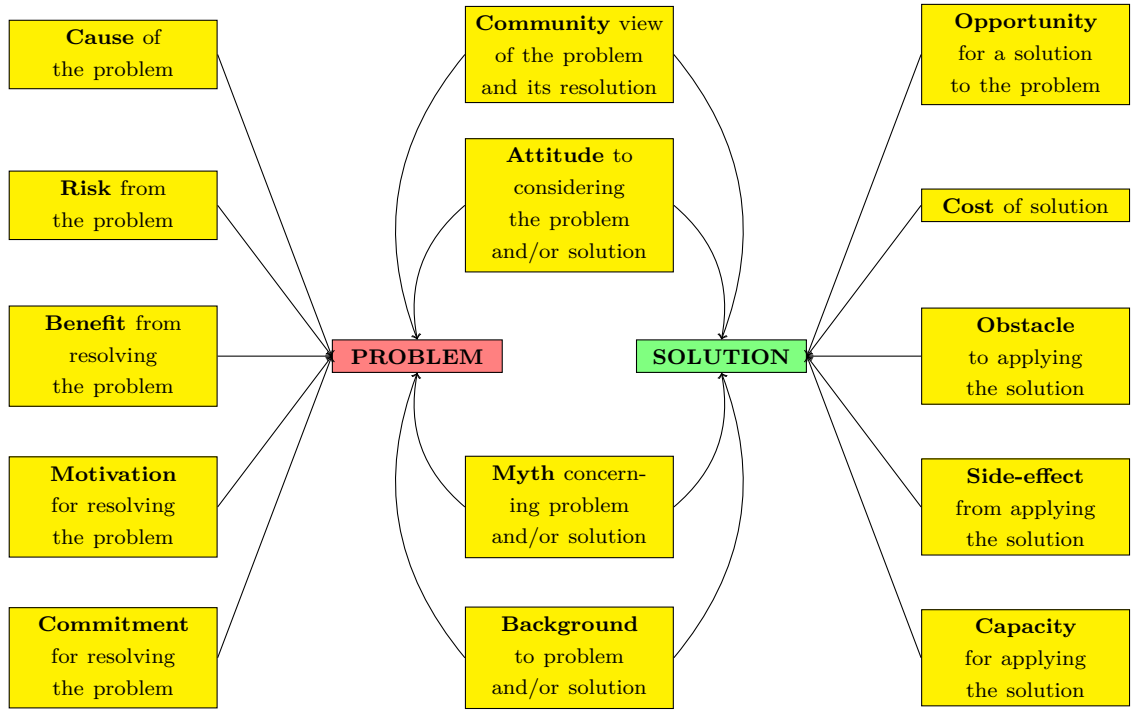

\begin{itemize}

\item A {\bf cause} argument is an argument that gives facts or an opinion about the cause of a healthcare problem.
For example, an agent is having a persistent cough because she smokes.
As another example, an agent is overweight because she drinks too many soft drinks, or an agent is overweight because the food corporations make soft drinks too calorific.

\item An {\bf attitude} argument is an argument that gives an opinion on the agent's attitude to considering the problem and/or solution. For example, my doctor recommends that vaccine, and so I will take it, or I resent my doctor raising the question of my body weight, and so I refuse to consider suggestion for losing weight, or I switch off if I am given statistics.

\item A {\bf risk} argument is an argument that describes an event that can happen with some probability in a time frame with a negative payoff (e.g., smoking can result in lung cancer). Special cases are the following:
A {\bf large risk} (e.g., overdosing on heroin can cause death);
A {\bf small risk} (e.g., the malaria prophylactic chloroquine can cause stomach upsets);
A {\bf long-term risk} (e.g., smoking can result in lung cancer);
A {\bf short-term risk} (e.g., smoking can cause coughing); A {\bf high risk} (e.g., long-term heavy drinking is likely to cause liver damage);
A {\bf low risk} (e.g., side effects from paracetamol are rare but can include an allergic reaction, which can cause a rash and swelling);
and A {\bf threat} is a special case where the system (or system owner) can invoke a sanction on the user (e.g., texting while driving company vehicles will be treated as a serious disciplinary issue, or smoking on company premises will be treated as a serious disciplinary issue).
Risks can be ranked by the persuader by the perceived negative value to the persuadee. This negative value is an aggregate of the probability the risk will occur and the seriousness of that occurrence to the persuadee. For example, consider the risks $A$ and $B$ from sunbathing without suncream where $A$ is ``If I sunbathe without suncream, my skin will become wrinkled", and $B$ is ``If I sunbathe without suncream, I will get a melanoma". The persuader may think that the persuadee would regard $A$ as a bigger risk than $B$, and rank them accordingly. This ranking may then be used by the persuadee to select $A$ in preference to $B$ in a dialogue with the persuadee.

\item A {\bf benefit} argument describes an event that can happen with some probability in a time frame with a positive payoff (e.g., stopping smoking can result in better life expectancy).
Some benefits arguments may be uncertain or even incorrect.
For example, a long-term smoker might think that having smoked for so long, there is no healthcare benefit of giving up smoking.
As another example, in trying to persuade a teenager with sensitive skin to use sun protection cream, the teenager might believe that her current usage of factor 10 cream is sufficient.
Special cases of benefits arguments are short-term benefit, long-term benefit, etc. A {\bf reward} or {\bf incentive} is a special case where the system (or system owner) can invoke a payoff to the user in exchange for accepting a persuasion goal.
Benefits can be ranked by the persuader by the perceived positive value to the persuadee. This positive value is an aggregate of the probability the benefit will occur and the how good that occurrence will be to the persuadee. For example, consider the benefits $A$ and $B$ from ceasing to smoke where $A$ is ``If I give up smoking, my skin will look better", and $B$ is ``If I give up smoking, I am less likely to get lung cancer". The persuader may think that the persuadee would regard $A$ as a bigger benefit than $B$, and rank them accordingly. This ranking may then be used by the persuadee to select $A$ in preference to $B$ in a dialogue with the persuadee.

\item An {\bf opportunity} argument is an argument that provides facts or an opinion concerning an opportunity for resolving the healthcare problem. For example, for an agent who has an alcohol consumption problem, an option for attempting to resolve the problem is a self-help group for encouraging drinking in moderation. As another example, if someone is moving from the city to the countryside, then she is more likely to go for a walk.

\item A {\bf cost} argument is an argument that gives facts or an opinion concerning the cost of resolving the healthcare problem.
For example, the cost of membership for a local gym could be given in terms of the monthly fee (60 British Pounds per month) or it could be given as less than the price of a coffee each day. As another example, an unemployed person might believe that she cannot participate in exercise activities at her local sports centre because she lacks money, and so she might be unaware that often sports centres offer reduced cost or free activities for unemployed people. 

\item A {\bf commitment} argument is an argument that captures a commitment made by the agent on something related to the healthcare problem. For example, an agent may have committed to participate in a marathon next year. In this example, the healthcare problem may that the agent is unfit, the agent has made this commitment, and the agent needs to be persuaded to train for the marathon.


\item A {\bf motivation} argument is an argument that gives an opinion on the motivation of the agent for resolving the healthcare problem. For example, a user might forward an argument that she wants to get fit for her summer holiday, but she is reluctant to do anything in her free time other than watch television.

\item A {\bf capacity} argument is an argument that gives an opinion on the capacity of the agent for addressing the healthcare problem.
For example, if someone believes that she has a determined and tenacious character, then she should be able to complete an exercise course. As another example, in trying to persuade someone to increase the amount of exercise that she undertakes, an argument against her signing up for an exercise class is that she has signed up a number of times before and dropped out within a couple of weeks.

\item An {\bf obstacle} argument is an argument that gives an opinion on the obstacles that the agent might face in addressing the healthcare problem. For example, for someone who lives far from the nearest gym, the need to access the gym by driving a car might be an obstacle if she does not know how to drive.



\item A {\bf background} argument is an argument that gives additional points about the background to the problem or solution. For example, NHS staff want the best for their patients, or people who do sports come in all shapes and sizes. A background argument is the category for any relevant argument does not fall into any of the other categories.

\item A {\bf side-effect} argument is an argument that gives facts or opinion on whether or not the solution has side-effects. For example, flu vaccines can cause an anaphylactic reaction, or pregnant women safely be given inactivated flu vaccine.

\item A {\bf community} argument is an argument that gives facts or an opinion on the cost to the community of the problem and/or of the benefit to the community of resolving the healthcare problem.
For example, the family of an agent might think that she drinks too much, and this has a negative impact on her health and behaviour, and that they would be happy if she reduced her consumption to a more healthy level.

\item A {\bf myth} argument is an is an argument argument that is based on information that is commonly held to be true but that is competely false with respect to the (scientific, legal, administrative, etc.) facts. For example, lung cancer is a male, working class disease.
We regard erroneous arguments as requiring a special category which we call myth. When modelling a domain, we want to clearly flag these arguments so that it is clear there is no doubt about them. Furthermore, when an erroneous argument is used in a dialogue, the system needs a strategy that is careful to strongly argue against it. In the field of health communications, such erroneous arguments would be described as contested health information that involves controversy, and mistrust. Frequently, it is based on some form of alleged conspiracy and spreads inaccurate, incorrect, inappropriate information \cite{Suggs15}.

\end{itemize}

We provide a summary of the ontology in Table \ref{table:ontology} where we give the correspondence with behaviour change modelling as discussed in Section \ref{section:background}.

\begin{table}
\centering
\begin{tabular}{lll}
\toprule
Type & Concerns & Correspondence to behaviour change modelling\\
\toprule
Cause & P & Belief concerning causes of the problem \\
Attitude & P/S & N/A \\
Risk & P & Belief concerning risks from the problem \\
Benefit & P & Belief concerning benefits from resolving the problem\\
Opportunity & S & Belief concerning opportunities for resolving the problem \\
Cost & S & Belief concerning costs of resolving the problem \\
Commitment & P/S & Belief concerning commitments of agent\\
Motivation & P & Belief concerning agent's motivation for addressing the problem \\
Capacity & S & Belief concerning agent's capacity for addressing the problem \\
Obstacle & S & Belief concerning obstacles for resolving the problem\\
Background & P/S & N/A \\
Side-effect & S & N/A \\
Community & P/S & Beliefs concerning the agent's community\\
Myth & P/S & N/A \\
\bottomrule
\end{tabular}
\caption{Summary of ontology\label{table:ontology}. For each type, we give whether it concerns the problem (denoted P) or the solution (S), and how it related to the behavioural change models we discussed in Section \ref{section:background}.}
\end{table}

\subsection{Functional types}
\label{section:functional}

A functional type for an argument delineates the function that the argument can have in a dialogue.
Arguments presented in a dialogue are taken from the domain model.
Domain model contains arguments of the following functional types, and these constitute the basic disjoint categories of argument for the purposes of computational persuasion in behaviour change.

\begin{description}

\item[Prospective] This is an argument that is a goal and therefore reflects how a persuader or persuadee would like the world to be. For example, the persuadee might have a goal that she would like to lose weight, or the persuader might have the goal to have more people taking the flu vaccine.

\item[Objective] This is an argument that is based on information that we can regard (from the point of view of the developers of the behaviour change application) as incontrovertible in the context of the behaviour change application. It could be based on well-established scientific or medical knowledge, or on current regulations and healthcare guidelines. For example, pregnant women can safely be given inactivated flu vaccine.

\item[Subjective] This is an argument that is based on information that we can regard (from the point of view of the developers of the behaviour change application) as controvertible in the context of the behaviour change application. It could be based on less established scientific or medical knowledge, on opinions, and attitudes. For example, flu is not a serious illness. This category also includes an argument that is based on information that we can regard (from the point of view of the developers of the behaviour change application) as completely incorrect in the context of the behaviour change application. For example, needless use of vaccines increases vaccine-resistant pathogens. We refer to such an erroneous argument as a myth.

\end{description}

The above categories are consistent with healthcare guidelines for managing people who have (potential) healthcare issues that would benefit from a change of behaviour. Consider, for example, the UK NICE {\em Guideline on Weight Management: lifestyle services for overweight and obese} \cite{NICE2014}. In recommendation 7, the guideline advises ``Discuss realistic weight-loss goals'' (which involves arguments concerning goals) and ``Discuss the importance and wider benefits of adults who are overweight or obese making gradual, long-term changes to their dietary habits and physical activity levels'' (which involves arguments arguments concerning objective and subjective information).





\begin{figure}
\begin{center}
\begin{tikzpicture}[->,scale=0.8]
\node (g1) [shape=rectangle,fill=yellow,draw] at (10,18) {Prospective};
\node (g2) [shape=rectangle,fill=yellow,draw] at (7,21) {Persuasion goal};
\node (g3) [shape=rectangle,fill=yellow,draw] at (10,21) {User goal};
\node (g4) [shape=rectangle,fill=yellow,draw] at (13,21) {Societal goal};
\path[]	(g2) edge[] (g1);
\path[]	(g3) edge[] (g1);
\path[]	(g4) edge[] (g1);
\node (i1) [shape=rectangle,fill=yellow,draw] at (5,16) {Functional};
\node (i2) [shape=rectangle,fill=yellow,draw] at (0,18) {Objective};
\node (i3) [shape=rectangle,fill=yellow,draw] at (5,14) {Subjective};
\path[]	(g1) edge[] (i1);
\path[]	(i2) edge[] (i1);
\path[]	(i3) edge[] (i1);
\node (o1) [shape=rectangle,fill=yellow,draw] at (-3,21) {Factual};
\node (o2) [shape=rectangle,fill=yellow,draw] at (0,21) {Evidence};
\node (o3) [shape=rectangle,fill=yellow,draw] at (3,21) {Example};

\path[]	(o1) edge[] (i2);
\path[]	(o2) edge[] (i2);
\path[]	(o3) edge[] (i2);

\node (s1) [shape=rectangle,fill=yellow,draw] at (1,12) {Preference};
\node (s4) [shape=rectangle,fill=yellow,draw] at (5,12) {Opinion};
\node (s9) [shape=rectangle,fill=yellow,draw] at (9,12) {Counter-bias};

\path[]	(s1) edge[] (i3);
\path[]	(s9) edge[] (i3);
\path[]	(s4) edge[] (i3);
\end{tikzpicture}
\end{center}
\caption{\label{fig:types}Graphical representation of subtypes of the functional type.}
\end{figure}
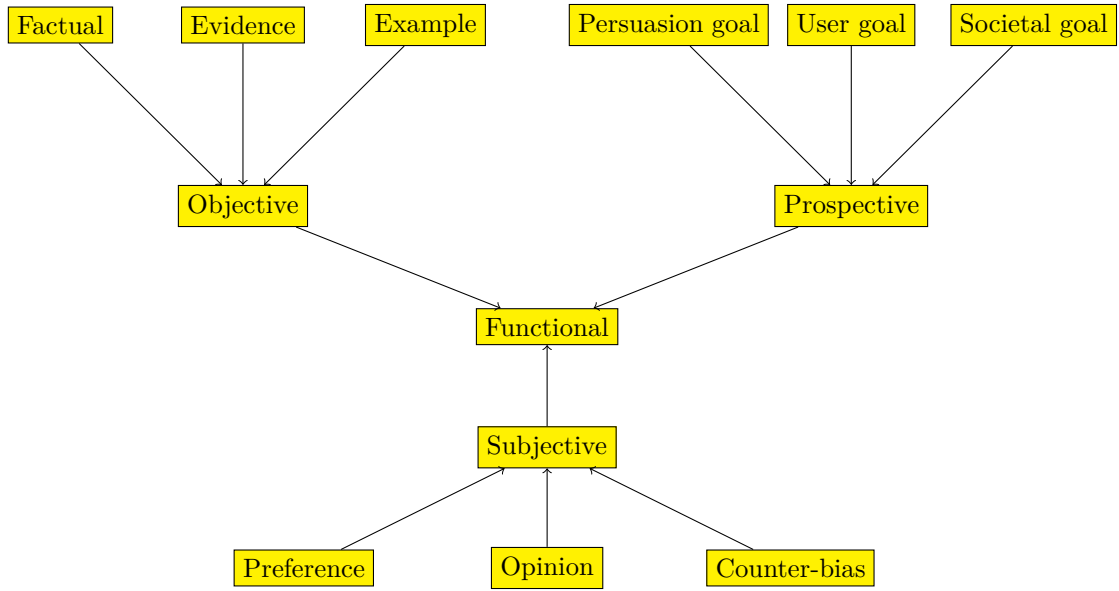

\subsubsection{Prospective arguments}

A goal is something that one of the agents (i.e., the persuadeer or persuadee) wants to be true, and the goal concerns the persuadee.
It may specify a desired goal state or action(s).
For instance, a persuader may have a goal specified as one or more of the following:
(1) A desired goal state (e.g., lose 10Kg);
(2) A desired action (e.g., book an appointment with a counselor);
and
(3) A desired sequence of actions (e.g., go to the gym every Tuesday and Thursday).

\begin{itemize}

\item A {\bf persuasion goal} argument describes a goal that the system wants to persuade the user to accept (e.g., do 2km walk every day). It may be to do something (e.g., to join an exercise class) or to not do something (e.g., to stop smoking). The system wants the user to believe a persuasion goal, and to be convinced by it.

\item A {\bf user goal} argument describes a specific goal that the user may have. Special cases are short-term goal (e.g., be fit for village fun run), long-term goal or desire (e.g., to live a long healthy life), etc.

\item A {\bf societal goal} argument describes a general goal that would normally be incontrovertible for any normal user (e.g., not waste government money).

\end{itemize}

A set of goal arguments is not necessarily consistent for a particular persuadee. For example, a persuasion goal might be ``I need to eat more healthly, therefore I should eat less chocolate'', and a user goal might be ``I need to relax more, therefore I should eat more chocolate''. For a persuadee who needs to eat more healthy and needs to relax more, there is a contradiction concerning whether or not to eat more chocolate.

Goal arguments can be ranked by desirability. By ranking goals, we can aim to get the highest ranked believed. If we fail, then we aim to get next highest believed, etc.

\begin{itemize}
\item Persuasion goal can by ranked by utility to the system (e.g., 2km walk per day is better than 10km walk per day from the user's point of view)
\item User goal can by ranked by utility to the user (e.g., 30min exercise class is better than 90min exercise class)
\item Societal goal can by ranked by utility to society (e.g., taking public transport is better than taking the car)
\end{itemize}

Ranking of goals is valuable extra information for the persuader in a dialogue. For instance, in a dialogue where the persuasion goal A = ``You could walk 10km daily to improve your health'' is presented by the system, and the user expresses strong agreement with the counterargument B = ``I have never walked 10km ever. It would be impossible.'' It is unlikely that the system would have a counterargument B that would cause her to believe A. The system could give up on A, and present a weaker persuasion goal such as C = ``You could walk 2km daily to improve your health".

To support the ranking of persuasion goals, we assume that the there is a ranking relation over persuasion goals, and that this is defined from the point of view of the persuader.
For example, suppose we have the arguments
$A_1$ = ``You could walk 10KM daily to improve your health'',
$A_2$ = ``You could walk 5KM daily to improve your health'',
$A_3$ = ``You could walk 2KM daily to improve your health'',
and $A_4$ = ``You could walk 1KM daily to improve your health''.
Then we could assume a ranking
where $A_1$ is ranked 1,
$A_2$ is ranked 2,
$A_3$ is ranked 3,
and $A_4$ is ranked 4.



\subsubsection{Objective arguments}
\label{section:objective}

An {\bf objective} argument is based on objective information which is information that is incontrovertible or information with substantial information to support its veracity such scientific knowledge, and statistical data. Special cases are definition, regulation, established scientific knowledge, utilitarian, user fact, etc.

\begin{itemize}

\item A {\bf factual} argument is an item of information that is normally incontrovertible such as definitions (e.g., alcopops are a kind of alcoholic drink), regulations (e.g., it is illegal to text while driving), established scientific knowledge (e.g., around half of all life-long smokers will die early), and user facts (e.g., the persuadee is middle-aged).

\item An {\bf evidence} argument is a source of information to support a factual such as a link to a web page. It provides more detailed information to reinforce a factual argument. It is possible that there may be a chain of evidence arguments where each provides increasingly detailed information in support of the argument at the start of the chain.

\item An {\bf example} argument is an instance of fact that supports the fact by being an illustration. Sometimes an example can communicate a lot of information. For instance, if someone has difficulty seeing how she could take up a sport, examples could be given of individuals similar to the persuadee, and these examples would show that it is indeed in possible. Furthermore, examples can be easier to understand than more general or abstract information.

\end{itemize}

Objective arguments are predominantly deployed by the system. They are the vehicle by which key items of information are presented to the user in order to persuade her of the value of agreeing to the persuasion goal, and of demonstrating the veracity of any argument supporting the persuasion goal.

\subsubsection{Subjective arguments}
\label{section:subjective}

A {\bf subjective} argument is based on subjective information. It is a kind of argument that the system can choose to query the user about to ask for the users degree of belief or agreement with the argument. Special cases are opinion, and preference.

\begin{itemize}

\item An {\bf opinion} argument is a view on knowledge that may be controversial or that at least the persuadee may doubt. Any of the categories presented in the ontology in Figure \ref{fig:ontology} may be the subject of an opinion. For example, a user may have the opinion that she has high motivation for exercise because she has signed up for numerous exercise classes, and the system may have an opinion that the user has a low motivation for doing exercise because every time she signs up for a new exercise class, she drops out after a week or two.

\item A {\bf preference} argument describes one or more preferences that the user has
 (e.g., I prefer take-away food to cooking at home).

 \item A {\bf counter-bias} argument is an argument that highlights and counters a cognitive bias in the agent concerning the evaluation of themselves or of the healthcare problem. For example, an agent may be unrealistically optimistic about her illness susceptibility, and if so, this could be explained to the agent. See \cite{Weinstein87} for more discussion of optimism about illness susceptibility, and how this can be addressed through explanation.

\end{itemize}

Subjective arguments are predominantly arguments that reflect the persuadee's position. The system may ask the user about her belief or agreement in specific subjective, or the system may guess which ones the user believes or agrees with. Using this information, the system can then attempt to select appropriate counterarguments.

\subsection{Context type}

Some arguments only apply to some intended persuadees (i.e., some arguments are only applicable in some contexts), and we call these {\bf contextual arguments}. For example, some arguments might only be applicable to people who live in London, or to people under the age of 18.
Contextual arguments would often be acquired from websites of local service providers (e.g., sports and leisure centres, local transport operators, social services, community centres, local health organizations, etc.).

In order to ensure that only applicable arguments are presented to persuadees, we can tag them according to context.
Some key dimensions for context types for contextual arguments are the following.

\begin{itemize}

\item {\bf Age context} This specifies the age range of a user in which an argument is applicable. For example, ``you can get to the country for a long walk using the bus service for free'' is an argument that might only apply to people greater than 64 years of age.

\item {\bf Geographical context} This specifies the geographical area of a user for which an argument is applicable. For example, ``you should take your medicine as prescriptions are free'' is an argument that applies to a user in Scotland but not in England.

\item {\bf Status context} This specifies the status the persuadee has to have for an argument to be applicable. For example, ``there is a free yoga class for unemployed people at your local gym''"'' is only applicable to unemployed people.

\end{itemize}

Context types would need to be tagged by annotators who have the background knowledge concerning the applicable of the arguments. Once tagged by context type, an argument can only be used in a dialogue if the model of the persuadee satisfies the context tag. For example, for the argument ``you should take your medicine as prescriptions are free'', if the user model contains the information that the persuadee lives in Scotland, then the argument can be used in the dialogue, whereas if the user model contains the information that the persuadee lives in England, then the argument cannot be used in the dialogue.

In order to automate the use of the context type, logic programming or a description logic ontology could be used. For example, a tagging of an argument for the geographical context being restricted to England could be captured by a literal of the form ${\tt geo(England)}$, and in the user model, the fact that the user lives in London captured by a literal of the form ${\tt geo(London)}$. Then, a rule stating that ${\tt geo(London)}$ implies ${\tt geo(England)}$ (i.e., ${\tt geo(London)} \Rightarrow {\tt geo(England)}$) could be used to infer that the geographical contraint of the argument is satisfied by the user.

The use of context types is consistent with healthcare guidelines for managing (potential) healthcare problems. Consider, for example, the UK NICE {\em Guideline on Weight Management: lifestyle services for overweight and obese} \cite{NICE2014}. In recommendation 8, the guideline advises to ``discuss with adults considering a lifestyle weight management programme what the programme does and does not involve, ....,
other local services that may provide additional support (for example, local walking or gardening groups), any financial costs (including any costs once a funded referral has ended)'', and in addition, it advises to ``explore ... any concerns they may have, or barriers they may face, in relation to joining the programme, the process of change or meeting their personal goals''. These recommendations therefore call for consideration of the context of the user, and adaptation of the dialogue accordingly.

\subsection{Topic types}

Since some agents will be more interested in some topics than others, it is good to choose arguments that are more interesting to the agent.



Topic types can be assigned using automated methods such as from information retrieval and text classification. The simplest approach is to use keywords. Here the text of an argumentation is treated as a ``bag of words" (i.e., this is the set of words in the text minus so called ``stop words"). Stop words are simple and very common words such as articles, pronouns, conjunctions, and common verbs. Commonly used lists of stop words contain around 200 to 300 words. Yet, around half of word occurrences used in text are stop words. The remaining words are regarded as having some role in indicating the meaning of the text when text is treated as a set of words.
We illustrate keyword assignment in the example in Figure \ref{fig:topics}.


\begin{figure}
\[
\begin{array}{l}
\keys(pg) = \{\type{sports activity, sign-up} \}\\
\keys(a1) = \{\type{self-conscious, body, sports} \}\\
\keys(a2) = \{\type{intimidated, sport} \}\\
\keys(a3) = \{\type{sports clubs, intimidating, unwelcoming} \}\\
\keys(b1a) = \{\type{body, sports} \}\\
\keys(b2a) = \{\type{clothes, sign-up} \}\\
\keys(b2b) = \{\type{sports} \}\\
\keys(c1a) = \{\type{sports, all shapes and sizes} \}\\
\keys(c2a) = \{\type{sports, care, sports fashion} \}\\
\keys(c3a) = \{\type{sports clubs, welcome new members, offer beginners classes} \}\\
\end{array}
\]
\caption{\label{fig:key}Consider Figure \ref{fig:topic}. For each argument $X$, $\keys(X)$ gives the keywords in $X$. }
\end{figure}

Since many arguments are likely to have keywords in common, these keywords in common are not good discriminators. Keywords that are less common in the graph can be good discriminators (e.g., in the above example, ``clothes" is a good discriminator that if expressed as an interest or concern by the user, would allow the system focus on this part of the argument graph). Standard text classification metrics can be used to automatically identify the best keywords to use for indexing arguments.

Another option is to use a fixed ontology for tagging the topics arising from each argument. The use of ontologies is wide-spread for cataloging content. Existing ontologies could be adapted for providing the range of topic indices that would be required for this role, or bespoke ontology could be developed for the domain of the app. Identifying the topic of an argument, offers a better way than keywords in helping to identify the arguments that may be of interest or important to a user. We illustrate the use of topic tagging in the example in Figure \ref{fig:topics}.

\begin{figure}
\[
\begin{array}{l}
\topics(pg) = \{\type{SportParticipation} \}\\
\topics(a1) = \{\type{SportParticipation, SelfConsciousAboutBody} \}\\
\topics(a2) = \{\type{TooIntimidated4Sport} \}\\
\topics(a3) = \{\type{SportsClubsIntimidating, SportsClubsUnwelcoming} \}\\
\topics(b1a) = \{\type{SportParticipation,Body4Sport} \}\\
\topics(b2a) = \{\type{Clothes4Sport} \}\\
\topics(b2b) = \{\type{Knowledge4Sport} \}\\
\topics(c1a) = \{\type{DiverseBodyShapeInSport} \}\\
\topics(c2a) = \{\type{SportsFashionUnimportant} \}\\
\topics(c3a) = \{\type{SportsClubWelcomeNewMembers, SportsClubBeginnersClasses} \}\\
\end{array}
\]
\caption{\label{fig:topics}Consider Figure \ref{fig:topic}. For each argument $X$, $\topics(X)$ gives the topics in $X$.
}
\end{figure}


As well as using standard techniques for using keywords, and topic classification, of arguments, machine learning techniques could be used to train classifiers of content. While more challenging in terms of getting reliable classification, it may offer the ability to get classification corresponding to a more sophisticated ontology. For more information on automated techniques for information retrieval and text classification, see \cite{Manning2000,Bird2009}.

\subsection{Summary of types and their utility}

We have proposed four main dimensions to the typing of arguments in the domain model. We summarize these as follows.

\begin{itemize}

\item {\bf Ontological} This type describes the kind of content of the argument based on the key factors used in models of behaviour change as we discussed in Section \ref{section:background}.

\item {\bf Functional} This type describes the roles that an argument may take in a discussion. It is based on the kind of content in an argument. For instance, the system can choose to start a dialogue with a persuasion goal so that the user is clear from the start what the goal is, or the system can start with a user goal, so that the user is clear form the start how the dialogue relates to her agenda.


\item {\bf Context} This type labels the argument with information delineating the applicability of the argument. If a user falls outside the area of applicability, the argument is irrelevant, and so should not be used, as it does not apply to that user.


\item {\bf Topic} This type describes the topic of an argument. If the user expresses an interest in a topic, then arguments labelled with these topics are more likely to be of interest to the user.


\end{itemize}

These types form the basis of a methodology for domain modelling. In order to construct a domain model for an application, we can use the hierarchy of functional types (i.e., Figure \ref{fig:types}) as the basis of questions to investigate the domain, and thereby collate relevant arguments. For instance, we ask about what are the possible persuasion goals, user goals, or societal goal relevant to this application. Similarly, we can ask what are the self-perceptions, attitudes to social norms, attitudes to knowledge, etc. Once the arguments have been collated, they can be arranged as an argument graph where each node is an argument, and each arc denotes an attack or support relationship. In addition, all the types can be used to add meta-level labels to each argument in order to describe the arguments.

Note, we do not assume any constraints, in general, on the arcs between the arguments of various types. For instance, it is possible to have two arguments $A$ and $B$ where both are persuasion goals, and $A$ attacks $B$ and $B$ attacks $A$. Similarly, we can have user goals that attack each other, and we can have a user goal attacking a persuasion goal and vice versa.
For example, it is possible for an agent to have the goal of being fit and to have the goal of doing no exercise.
However, some kinds of attack would be unusual for a behaviour change application. For instance, it would be unusual for an argument that is a factual type to attack another argument that is also of a factual type. As another example, we would expect that an erroneous argument or myth argument is attacked by a factual argument.

In the case studies, presented in Section \ref{section:casestudies}, we provide four examples of labelling with ontological types and discuss the the functional, context topic types for them. Then in Section \ref{section:use}, we discuss how the user model could be designed to take advantage of the domain modelling.



\section{Case studies}
 \label{section:casestudies}

\begin{figure}





\begin{center}
\begin{tikzpicture}[->,>=latex,thick,scale=0.5, every node/.style={scale=0.6}]


\node (pg) [text centered,text width=3.5cm,shape=rectangle,fill=red!50,draw] at (12,20) {[Persuasion goal] I will go for a smear test soon as I haven't been for at least 3 years.};


\node (s1) [text centered,text width=3.5cm,shape=rectangle,fill=yellow,draw] at (10,32) {[s1 Obstacle] I am embarrassed to at exposing private parts of my body to a stranger.};

\node (s2) [text centered,text width=2.5cm,shape=rectangle,fill=yellow,draw] at (15,28) {[s2  Obstacle]  I fear medical procedures.};

\node (s3) [text centered,text width=2.5cm,shape=rectangle,fill=yellow,draw] at (20,28) {[s3 Obstacle] I fear the pain.};

\node (s4) [text centered,text width=2.5cm,shape=rectangle,fill=yellow,draw] at (25,28) {[s4  Obstacle] I don't like hospital or doctors surgeries.};

\path[]	(s1) edge[] node[] {} (pg);
\path[]	(s2) edge[] node[] {} (pg);
\path[]	(s3) edge[] node[] {} (pg);
\path[]	(s4) edge[] node[] {} (pg);


\node (r1) [text centered,text width=6cm,shape=rectangle,fill=yellow,draw] at (10,36) {[r1 Opportunity] Healthcare professionals are experienced in examining patients, and are able to put you at ease about the examination.};

\path[]	(r1) edge[] node[] {} (s1);

\node (r2) [text centered,text width=2.5cm,shape=rectangle,fill=yellow,draw] at (15,32) {[r2  Opportunity] It is just a simple non-painful procedure.};

\path[]	(r2) edge[] node[] {} (s2);

\node (r4) [text centered,text width=2.5cm,shape=rectangle,fill=yellow,draw] at (20,32) {[r4  Opportunity] It is something that is done very quickly.};

\path[]	(r4) edge[] node[] {} (s3);


\node (h1) [text centered,text width=2.5cm,shape=rectangle,fill=yellow,draw] at (6,28) {[h1 Risk] I'm not sexually active, and so I don't need to do it.};


\node (i1) [text centered,text width=2.5cm,shape=rectangle,fill=blue!20,draw] at (0,28) {[i1 Risk] If you have ever been sexually active, then the medical advice is to participate in screening.};

\path[]	(h1) edge[] node[] {} (pg);
\path[]	(i1) edge[] node[] {} (h1);

\node (h2) [text centered,text width=2.5cm,shape=rectangle,fill=yellow,draw] at (6,22) {[h2  Risk] I've had a partial hysterectomy, and so I don't need to do it.};

\node (i2) [text centered,text width=2.5cm,shape=rectangle,fill=blue!20,draw] at (0,22) {[h2 Risk] Normally, the medical advice is to continue with screening.};

\path[]	(h2) edge[] node[] {} (pg);
\path[]	(i2) edge[] node[] {} (h2);


\node (m1) [text centered,text width=2.5cm,shape=rectangle,fill=yellow,draw] at (18,20) {[m1 Obstacle] I don't have enough time to do the test.};

\path[]	(m1) edge[] node[] {} (pg);

\node (j1) [text centered,text width=3.5cm,shape=rectangle,fill=yellow,draw] at (24,22) {[j1 Risk] Think how you would feel if you only found out that you had it when had advanced.};

\path[]	(j1) edge[] node[] {} (m1);
\path[]	(j1) edge[] node[] {} (s4);

\node (j2) [text centered,text width=3.5cm,shape=rectangle,fill=yellow,draw] at (24,18) {[j2 Benefits] You will feel guilty if you don't do the test.};

\path[]	(j2) edge[] node[] {} (m1);


\node (a0) [text centered,text width=2.5cm,shape=rectangle,fill=yellow,draw] at (0,16) {[a0  Community] Many of my friends do it.};

\node (a1) [text centered,text width=2.5cm,shape=rectangle,fill=yellow,draw] at (0,12) {[a1  Risk] I am scared of getting cancer.};

\node (a2) [text centered,text width=2.5cm,shape=rectangle,fill=yellow,draw] at (5,12) {[a2 Attitude] Since my doctor recommends it, I will do it.};

\node (a3) [text centered,text width=2.5cm,shape=rectangle,fill=yellow,draw] at (10,12) {[a3 Benefit] The benefit in terms of piece of mind outweighs the hassle of doing it.};

\node (a4) [text centered,text width=2.5cm,shape=rectangle,fill=yellow,draw] at (15,12) {[a4 Risk] The absolute risk of getting cervical cancer is very low.};

\node (a5) [text centered,text width=2.5cm,shape=rectangle,fill=yellow,draw] at (20,12) {[a5 Risk] The test can have false positives.};

\node (a6) [text centered,text width=2.5cm,shape=rectangle,fill=yellow,draw] at (25,12) {[a6 Obstacle] I don't know if I could cope with getting a positive result.};

\path[]	(a0) edge[dashed] node[] {} (pg);
\path[]	(a1) edge[dashed] node[] {} (pg);
\path[]	(a2) edge[dashed] node[] {} (pg);
\path[]	(a3) edge[dashed] node[] {} (pg);
\path[]	(a4) edge[] node[] {} (pg);
\path[]	(a5) edge[] node[] {} (pg);
\path[]	(a5) edge[] node[] {} (a6);
\path[]	(a6) edge[] node[] {} (pg);


\node (b4) [text centered,text width=3.5cm,shape=rectangle,fill=yellow,draw] at (12,6) {[b4 Benefit] The benefit of getting the test done greatly outweighs the cost and the low absolute risk.};

\path[]	(b4) edge[dashed] node[] {} (a4);
\path[]	(b4) edge[dashed] node[] {} (a3);

\node (b5) [text centered,text width=4.5cm,shape=rectangle,fill=yellow,draw] at (22,6) {[b5 Benefits] The proportion of false positives are relatively small, and normally they can be quickly identified with some further tests.};

\path[]	(b5) edge[] node[] {} (a5);
\path[]	(b5) edge[] node[] {} (a6);


\end{tikzpicture}
\end{center}

\caption{\label{fig:smeargraph}Argument graph for participation in cervical smear screening. A dashed line denotes a support and a solid line denotes an attack. The problem is {\em risk of cervical cancer} and the proposed solution is {\em participate in screening programme}.}
\end{figure}
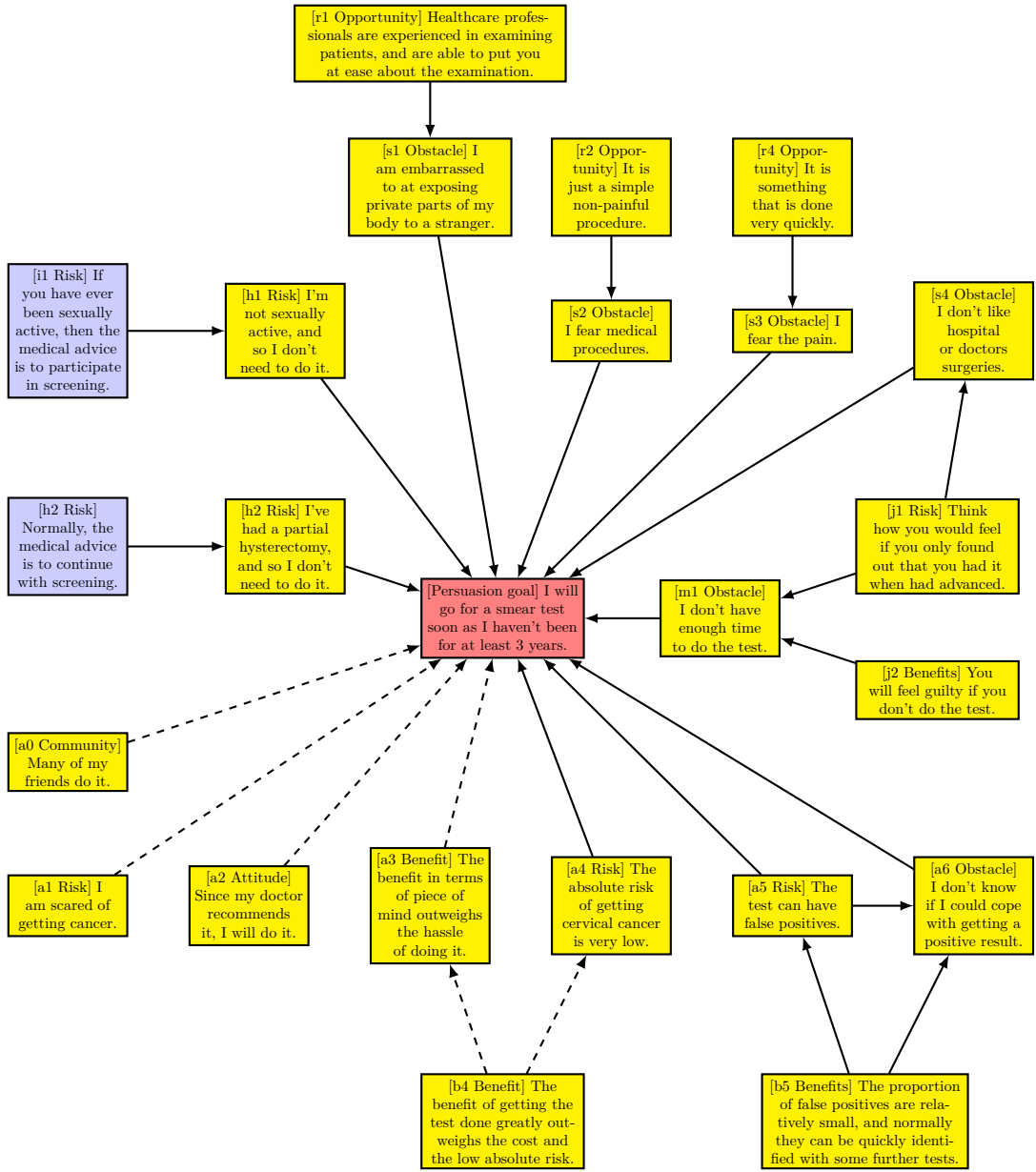

\begin{figure}
\input{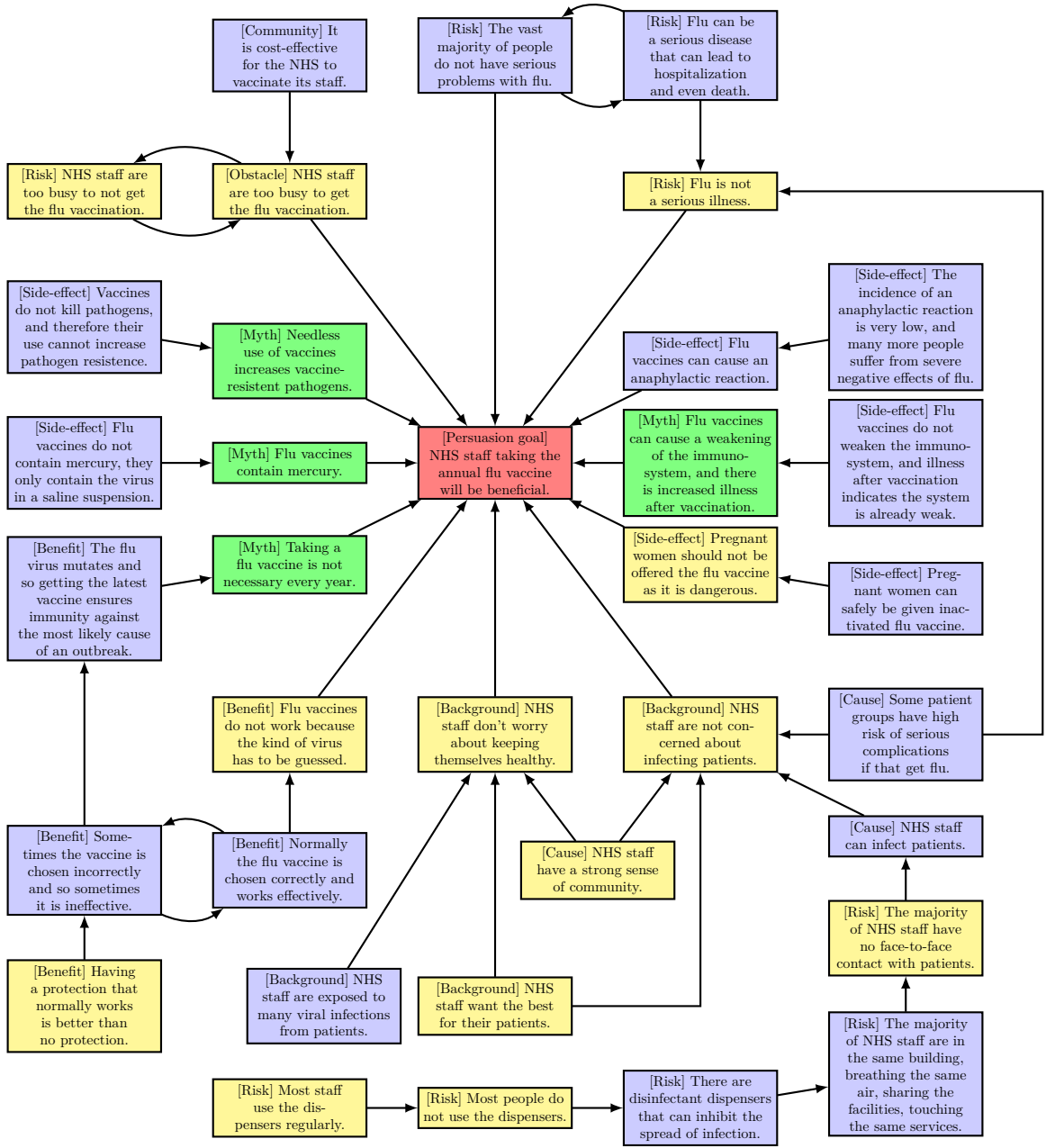}
\caption{\label{fig:vaccinegraph2}Argument graph for flu vaccine for healthcare workers. A dashed line denotes a support and a solid line denotes an attack. The problem is {\em risk of influenza in healthcare workers} and the proposed solution is {\em participate in vaccination programme}.}
\end{figure}

In order to test the dimensions for domain modelling, we undertake four case studies. Each case study concerns an important topic in behaviour change. For each case study, we produced an argument graph where each argument is labelled with a functional type. We briefly describe each case study below, including some key sources we used for identifying relevant arguments, and we give the argument graphs in Figures \ref{fig:sportgraph} to \ref{fig:vaccinegraph2}. Note, for each argument graph, we aim to give a non-exhaustive variety of arguments that could be used in computational persuasion. Based on the resources we have available, it would be straightforward to expand each graph with further arguments.


\begin{description}

\item[Women in sport]
We based this case study on a Sports England study into the engagement of young women in sport
which was commissioned to understand reasons for participating and for not participating \cite{Cox2006}.
Further arguments come from a Sports England initiative, This Girl Can, for encouraging young women into sports participation\footnote{http://www.thisgirlcan.co.uk/feel-inspired/}, and the biomedical literature \cite{Kilpatrick05,Allison05}. The argument graph for this case study is given in Figure \ref{fig:sportgraph}. The healthcare problem being addressed is {\em insufficient exercise} and the solution is {\em participate in sport}.

\item[Office wellbeing initiative]
There is increasing interest from organizations in organizing resources and initiatives to help their employees improve their wellbeing. One kind of activity is the formation of groups in an office for specific regular exercise such as running or walking 10km each week. Sources for arguments include a report from the Institute for Occupational Health on health and wellbeing in the workplace \cite{IOSH15}, a report from the Scottish Executive Social Research \cite{Murray06} on participation in sport, information from the Workplace Wellbeing Charter website\footnote{http://www.wellbeingcharter.org.uk} on workplace exercise initiatives, and the biomedical literature \cite{White16} on workplace exercise initiatives. The argument graph for this case study is given in Figure \ref{fig:officegraph}. The healthcare problem being addressed is {\em insufficient exercise} and the solution is {\em participate in office initiative}.

\item[Cervical screening]
Cervical screening, also known as the smear test, is a method to identify abnormal cell growth on the cervix. Since these cells may lead to cervical cancer, and it can be advised that the growth is removed in a surgical procedure. Despite the value of cervical screening in identifying potential problems early, many women do not participant in cervical screening programmes for various reasons.
Sources for arguments include the biomedical literature \cite{Harlan91,Eaker01,Urrutia17} and healthcare provider websites such as NHS Choices\footnote{http://www.nhs.uk/pages/home.aspx} and the Mayo Clinic\footnote{http://www.mayoclinic.org/tests-procedures/pap-smear/basics/why-its-done/prc-20013038}.
The argument graph for this case study is given in Figure \ref{fig:smeargraph}. The healthcare problem being addressed is {\em risk of cervical cancer} and the solution is {\em participate in screening programme}.

\item[Flu vaccine for healthcare workers]
In the UK, healthcare is provided by the National Health Service (NHS) which is a government-funded organization. It is a very large employer. For instance, NHS England has 1.3 million employees. With such a large number of works, there are many lost days of work caused by influenza. Furthermore, NHS employees may be vulnerable to infection from patients with influenza, and influenza passed onto patients causes complications. In order to address these problems, the NHS offers free flu vaccination to NHS employees in the workplace. Despite this, only around 50\% of NHS employees take the vaccine. Arguments for and against the flu vaccine were taken from the biomedical literature \cite{Marcua2015,Salisbury2013} and from websites that discuss benefits such as the Centers for Disease Control and Prevention\footnote{https://www.cdc.gov/flu/index.htm} and websites that discuss perceived problems with flu vaccinations such as the Mercola website\footnote{http://articles.mercola.com/sites/articles/archive/2015/12/01/another-flu-vaccine-flop.aspx}
and Natural Health 365\footnote{http://www.naturalhealth365.com/flu-shot-vaccine-dangers-1640.html}.
The argument graph for this case study is given in Figure \ref{fig:vaccinegraph2}. The healthcare problem being addressed is {\em risk of influenza} and the solution is {\em participate in vaccination programme}.

\end{description}


\begin{figure}





\begin{center}
\begin{tikzpicture}[->,>=latex,thick,scale=0.5, every node/.style={scale=0.6}]


\node (pg) [text centered,text width=3.5cm,shape=rectangle,fill=red!50,draw] at (12,20) {[Persuasion goal] I will sign up for a sports activity.};


\node (s1) [text centered,text width=2.5cm,shape=rectangle,fill=yellow,draw] at (0,28) {[s1 Benefit] Doing sport is a great way to relieve stress.};

\node (s2) [text centered,text width=2.5cm,shape=rectangle,fill=yellow,draw] at (6,28) {[s2 Benefit]  Doing sport is a great way to have fun.};

\node (s3) [text centered,text width=2.5cm,shape=rectangle,fill=yellow,draw] at (12,28) {[s3 Benefit] Doing sport is a great way of improving your social life.};

\node (s4) [text centered,text width=2.5cm,shape=rectangle,fill=blue!20,draw] at (18,28) {[s4 Benefit] Doing sport brings health benefits.};

\path[]	(s1) edge[dashed] node[] {} (pg);
\path[]	(s2) edge[dashed] node[] {} (pg);
\path[]	(s3) edge[dashed] node[] {} (pg);
\path[]	(s4) edge[dashed] node[] {} (pg);


\node (r2a) [text centered,text width=2.5cm,shape=rectangle,fill=yellow,draw] at (4,32) {[r2a Obstacle] I don't find sport is fun.};

\node (r2b) [text centered,text width=2.5cm,shape=rectangle,fill=yellow,draw] at (8,32) {[r2a Context] Sport at schools was not fun.};

\path[]	(r2a) edge[] node[] {} (s2);
\path[]	(r2b) edge[] node[] {} (s2);

\node (r3a) [text centered,text width=2.5cm,shape=rectangle,fill=yellow,draw] at (12,32) {[r3a Context] I already have plenty of friends.};

\path[]	(r3a) edge[] node[] {} (s3);

\node (r4a) [text centered,text width=2.5cm,shape=rectangle,fill=yellow,draw] at (18,32) {[r4a User goal] Getting fit is not important to me.};
\node (r4b) [text centered,text width=2.5cm,shape=rectangle,fill=yellow,draw] at (24,32) {[r4b Context] I am already fit.};

\path[]	(r4a) edge[] node[] {} (s4);
\path[]	(r4b) edge[] node[] {} (s4);


\node (q2a) [text centered,text width=3.5cm,shape=rectangle,fill=blue!20,draw] at (6,36) {[q2a Opportunity] There is a wide variety of fun sports activities to be tried from archery to zumba.};

\path[]	(q2a) edge[] node[] {} (r2a);
\path[]	(q2a) edge[] node[] {} (r2b);

\node (q3a) [text centered,text width=3.5cm,shape=rectangle,fill=yellow,draw] at (12,36) {[q3a Opportunity] There are some great ways you can do sports with your friends.};

\path[]	(q3a) edge[] node[] {} (r3a);

\node (q4a) [text centered,text width=3.5cm,shape=rectangle,fill=yellow,draw] at (20,36) {[q4a Benefit] When you get older, you will appreciate having got into the habit of doing exercise.};

\path[]	(q4a) edge[] node[] {} (r4a);
\path[]	(q4a) edge[] node[] {} (r4b);


\node (h1) [text centered,text width=2.5cm,shape=rectangle,fill=yellow,draw] at (6,20) {[h1 Obstacle] Sport is for males.};

\node (i1a) [text centered,text width=2.5cm,shape=rectangle,fill=blue!20,draw] at (0,18) {[i1a Context] There are many women competing at the highest level in many different sports.};

\node (i1b) [text centered,text width=2.5cm,shape=rectangle,fill=blue!20,draw] at (0,22) {[i1b Opportunity] Some sports clubs have activities specifically for women.};

\path[]	(h1) edge[] node[] {} (pg);
\path[]	(i1a) edge[] node[] {} (h1);
\path[]	(i1b) edge[] node[] {} (h1);


\node (m1) [text centered,text width=2.5cm,shape=rectangle,fill=yellow,draw] at (18,20) {[m1 Obstacle] I don't have enough time to do sport.};

\path[]	(m1) edge[] node[] {} (pg);

\node (j1a) [text centered,text width=2.5cm,shape=rectangle,fill=blue!20,draw] at (24,20) {[j1a Opportunity] There are sports activities that can be squeezed into any time of the day.};

\path[]	(s4) edge[] node[] {} (m1);
\path[]	(j1a) edge[] node[] {} (m1);


\node (a1) [text centered,text width=2.5cm,shape=rectangle,fill=yellow,draw] at (0,12) {[a1 Capacity] I feel too self-conscious about my body to do sport.};

\node (a2) [text centered,text width=2.5cm,shape=rectangle,fill=yellow,draw] at (6,12) {[a2 Capacity] I feel too intimidated to do sport.};

\node (a3) [text centered,text width=2.5cm,shape=rectangle,fill=yellow,draw] at (12,12) {[a3 Obstacle] Sports clubs seem intimidating and unwelcoming.};

\node (a4) [text centered,text width=2.5cm,shape=rectangle,fill=yellow,draw] at (18,16) {[a4 Obstacle] There is a lack of opportunities for doing sport.};

\node (a5) [text centered,text width=2.5cm,shape=rectangle,fill=yellow,draw] at (18,12) {[a5 Obstacle] There are hassles in order to participate in sport.};

\path[]	(a1) edge[] node[] {} (pg);
\path[]	(a2) edge[] node[] {} (pg);
\path[]	(a3) edge[] node[] {} (pg);
\path[]	(a4) edge[] node[] {} (pg);
\path[]	(a5) edge[] node[] {} (pg);


\node (b1a) [text centered,text width=2.5cm,shape=rectangle,fill=yellow,draw] at (0,6) {[b1a Obstacle] I do not have the body for sport.};

\path[]	(b1a) edge[dashed] node[] {} (a1);
\path[]	(b1a) edge[dashed] node[] {} (a2);

\node (b2a) [text centered,text width=2.5cm,shape=rectangle,fill=yellow,draw] at (6,6) {[b2a Obstacle] I do not have the right clothes to do sport.};

\node (b2b) [text centered,text width=2.5cm,shape=rectangle,fill=yellow,draw] at (10,6) {[b2b Obstacle] I do not know what I am supposed to do when I try to do sport.};

\path[]	(b2a) edge[dashed] node[] {} (a2);
\path[]	(b2b) edge[dashed] node[] {} (a2);

\node (b5a) [text centered,text width=2.5cm,shape=rectangle,fill=yellow,draw] at (16,7) {[b5a Example] Having to fill in forms.};

\node (b5b) [text centered,text width=2.5cm,shape=rectangle,fill=yellow,draw] at (20,7) {[b5b Example] Having enough money for fees.};

\node (b5c) [text centered,text width=2.5cm,shape=rectangle,fill=yellow,draw] at (24,7) {[b5c Example] Having to prove age.};

\path[]	(b5a) edge[dashed] node[] {} (a5);
\path[]	(b5b) edge[dashed] node[] {} (a5);
\path[]	(b5c) edge[dashed] node[] {} (a5);


\node (c1a) [text centered,text width=2.5cm,shape=rectangle,fill=yellow,draw] at (0,2) {[c1a  Background] People who do sport come in all shapes and sizes.};

\path[]	(c1a) edge[] node[] {} (b1a);

\node (c2a) [text centered,text width=2.5cm,shape=rectangle,fill=yellow,draw] at (6,2) {[c2a  Background] Most people who do sport really do not care about sports fashion.};

\path[]	(c2a) edge[] node[] {} (b2a);

\node (c3a) [text centered,text width=3.5cm,shape=rectangle,fill=blue!20,draw] at (12,2) {[c3a  Opportunity] Many sports clubs are very keen to welcome new comers, and offer beginners classes for them.};

\path[]	(c3a) edge[] node[] {} (b2b);
\path[]	(c3a) edge[] node[] {} (a3);

\node (c4a) [text centered,text width=3.5cm,shape=rectangle,fill=blue!20,draw] at (24,16) {[c4a  Opportunity] With the new government initiative, there are many opportunities to participate in women's sport in your area.};

\path[]	(c4a) edge[] node[] {} (a4);

\node (c5a) [text centered,text width=3.5cm,shape=rectangle,fill=blue!20,draw] at (20,2) {[c5a  Opportunity] With the new government initiative, you can join up for women's sport with the minimum of hassle.};

\path[]	(c5a) edge[] node[] {} (b5a);
\path[]	(c5a) edge[] node[] {} (b5b);
\path[]	(c5a) edge[] node[] {} (b5c);



\end{tikzpicture}
\end{center}

\caption{\label{fig:sportgraph}Argument graph for participation by women in sports. A dashed line denotes a support and a solid line denotes an attack. The problem is {\em insufficient exercise} and the proposed solution is {\em participate in sport}.}
\end{figure}
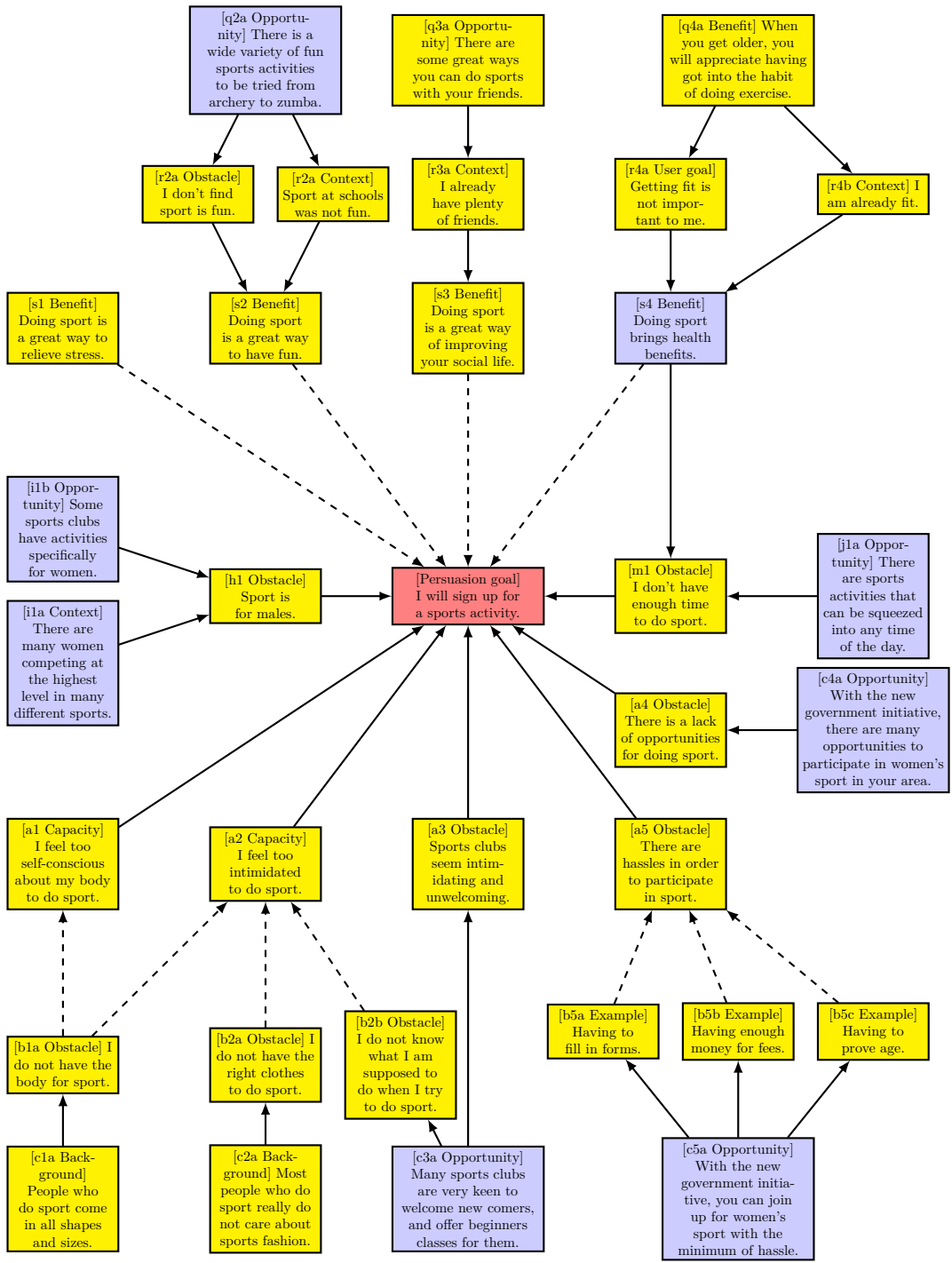

\begin{figure}





\begin{center}
\begin{tikzpicture}[->,>=latex,thick,scale=0.5, every node/.style={scale=0.6}]



\node (pg1) [text centered,text width=3.5cm,shape=rectangle,fill=red!50,draw] at (12,28) {[pg1 Persuasion goal] I will join the office group that walks 10km per week.};

\node (pr1) [text centered,text width=2cm,shape=rectangle,fill=pink!50,draw] at (10,22) {[pr1 Preference] I prefer walking to running.};

\node (pr2) [text centered,text width=2cm,shape=rectangle,fill=pink!50,draw] at (13.5,22) {[pr2 Preference] I prefer running to walking.};

\node (pg2) [text centered,text width=3.5cm,shape=rectangle,fill=red!50,draw] at (12,16) {[pg2 Persuasion goal] I will join the office group that runs 10km per week.};

\path[]	(pr1) edge[dashed] node[] {} (pg1);
\path[]	(pr1) edge[] node[] {} (pg2);
\path[]	(pr2) edge[] node[] {} (pg1);
\path[]	(pr2) edge[dashed] node[] {} (pg2);


\node (s1) [text centered,text width=3.5cm,shape=rectangle,fill=yellow,draw] at (24,28) {[s1  Benefit]  Doing it as an office group will be fun.};

\node (s2) [text centered,text width=3.5cm,shape=rectangle,fill=yellow,draw] at (24,24) {[s2 Benefit]  Signing up for an office group will make it harder to not do it regularly.};

\node (s3) [text centered,text width=3.5cm,shape=rectangle,fill=yellow,draw] at (24,20) {[s3  Benefit] Doing it as an office group is a great way of making friends.};

\node (s4) [text centered,text width=3.5cm,shape=rectangle,fill=yellow,draw] at (24,16) {[s4 Benefit] Doing regular exercise will be good for me.};

\path[]	(s1.west) edge[dashed] node[] {} (pg1);
\path[]	(s1.west) edge[dashed] node[] {} (pg2);
\path[]	(s2.west) edge[dashed] node[] {} (pg1);
\path[]	(s2.west) edge[dashed] node[] {} (pg2);
\path[]	(s3.west) edge[dashed] node[] {} (pg1);
\path[]	(s3.west) edge[dashed] node[] {} (pg2);
\path[]	(s4.west) edge[dashed] node[] {} (pg1);
\path[]	(s4.west) edge[dashed] node[] {} (pg2);




\node (h0) [text centered,text width=3.5cm,shape=rectangle,fill=yellow,draw] at (2,28) {[h1 Benefit] Walking does not require paying for any gear.};

\path[]	(h0) edge[] node[] {} (pg1);

\node (h1) [text centered,text width=2.5cm,shape=rectangle,fill=yellow,draw] at (6,22) {[h1  Obstacle] I don't have time for exercise.};

\node (i1a) [text centered,text width=2.5cm,shape=rectangle,fill=yellow,draw] at (0,22) {[i1a User goal] I don't want to appear anti-social in the office.};


\path[]	(h1) edge[] node[] {} (pg1);
\path[]	(h1) edge[] node[] {} (pg2);
\path[]	(i1a) edge[] node[] {} (h1);

\node (h2) [text centered,text width=2.5cm,shape=rectangle,fill=yellow,draw] at (5,16) {[h2  Benefit] You will feel proud if you can beat your colleagues at running.};

\path[]	(h2) edge[dashed] node[] {} (pg2);

\node (h3) [text centered,text width=2.5cm,shape=rectangle,fill=yellow,draw] at (0,16) {[h2 Background] I like to feel proud.};

\path[]	(h3) edge[dashed] node[] {} (h2);







\node (a0) [text centered,text width=2.5cm,shape=rectangle,fill=yellow,draw] at (0,10) {[a0 Obstacle] My colleagues will laugh at me if they see my legs in shorts.};

\node (a1) [text centered,text width=3.5cm,shape=rectangle,fill=yellow,draw] at (6,10) {[a1  Obstacle] I feel too self-conscious about my body to do sport.};

\node (a2) [text centered,text width=3.5cm,shape=rectangle,fill=yellow,draw] at (12,10) {[a2  Obstacle] I feel too intimidated to do sport.};

\node (a3) [text centered,text width=3.5cm,shape=rectangle,fill=yellow,draw] at (18,10) {[a3  Cost] Getting sweaty in front of my colleagues is too embarassing.};

\node (a4) [text centered,text width=2.5cm,shape=rectangle,fill=yellow,draw] at (24,10) {[a4  Risk] If I don't do regular exercise, I will have a raised risk of health problems.};

\path[]	(a0) edge[] node[] {} (pg2);
\path[]	(a1) edge[] node[] {} (pg2);
\path[]	(a2) edge[] node[] {} (pg2);
\path[]	(a3) edge[] node[] {} (pg2);
\path[]	(a4) edge[dashed] node[] {} (pg1);
\path[]	(a4) edge[dashed] node[] {} (pg2);


\node (b0) [text centered,text width=2.5cm,shape=rectangle,fill=yellow,draw] at (0,4) {[b0 Fact] Many runners wear pants rather than shorts.};

\path[]	(b0) edge[dashed] node[] {} (a0);

\node (b1a) [text centered,text width=3.5cm,shape=rectangle,fill=yellow,draw] at (6,6) {[b1a  Obstacle] I do not have the body for running.};

\path[]	(b1a) edge[dashed] node[] {} (a1);
\path[]	(b1a) edge[dashed] node[] {} (a2);

\node (b2a) [text centered,text width=3.5cm,shape=rectangle,fill=yellow,draw] at (12,6) {[b2a Obstacle] I do not have the right clothes to do running.};

\path[]	(b2a) edge[dashed] node[] {} (a2);

\node (b3) [text centered,text width=3.5cm,shape=rectangle,fill=yellow,draw] at (18,6) {[b3  Background] Everyone will be the same and so it really doesn't matter.};

\path[]	(b3) edge[] node[] {} (a3);


\node (c1a) [text centered,text width=3.5cm,shape=rectangle,fill=yellow,draw] at (6,2) {[c1a Background] People who do running come in all shapes and sizes.};

\path[]	(c1a) edge[] node[] {} (b1a);

\node (c2a) [text centered,text width=3.5cm,shape=rectangle,fill=yellow,draw] at (12,2) {[c2a Background] Most runners really do not care about sports fashion.};

\path[]	(c2a) edge[] node[] {} (b2a);

\node (c3) [text centered,text width=3.5cm,shape=rectangle,fill=violet!40,draw] at (24,2) {[c3 User goal] I want to stay healthy for my children / grandchildren.};

\path[]	(c3) edge[] node[] {} (a4);
\draw[dashed] (c3) -| (28,10) |- (s4.east);

\node (c4) [text centered,text width=3.5cm,shape=rectangle,fill=violet!40,draw] at (24,32) {[c3 User goal] I would like to make friends.};

\draw[dashed] (c4) -| (28,24) |-  (s3.east);   


\node (m1) [text centered,text width=3.5cm,shape=rectangle,fill=yellow,draw] at (12,32) {[m1 Motivation] I think walking is uninspiring.};

\path[]	(m1) edge[] node[] {} (pg1);

\path[]	(s1.west) edge[] node[] {} (m1.east);


\end{tikzpicture}
\end{center}

\caption{\label{fig:officegraph}Argument graph for office well-being initiative. A dashed line denotes a support and a solid line denotes an attack. The problem is {\em insufficient exercise} and the proposed solution is {\em participate in office initiative}.}
\end{figure}
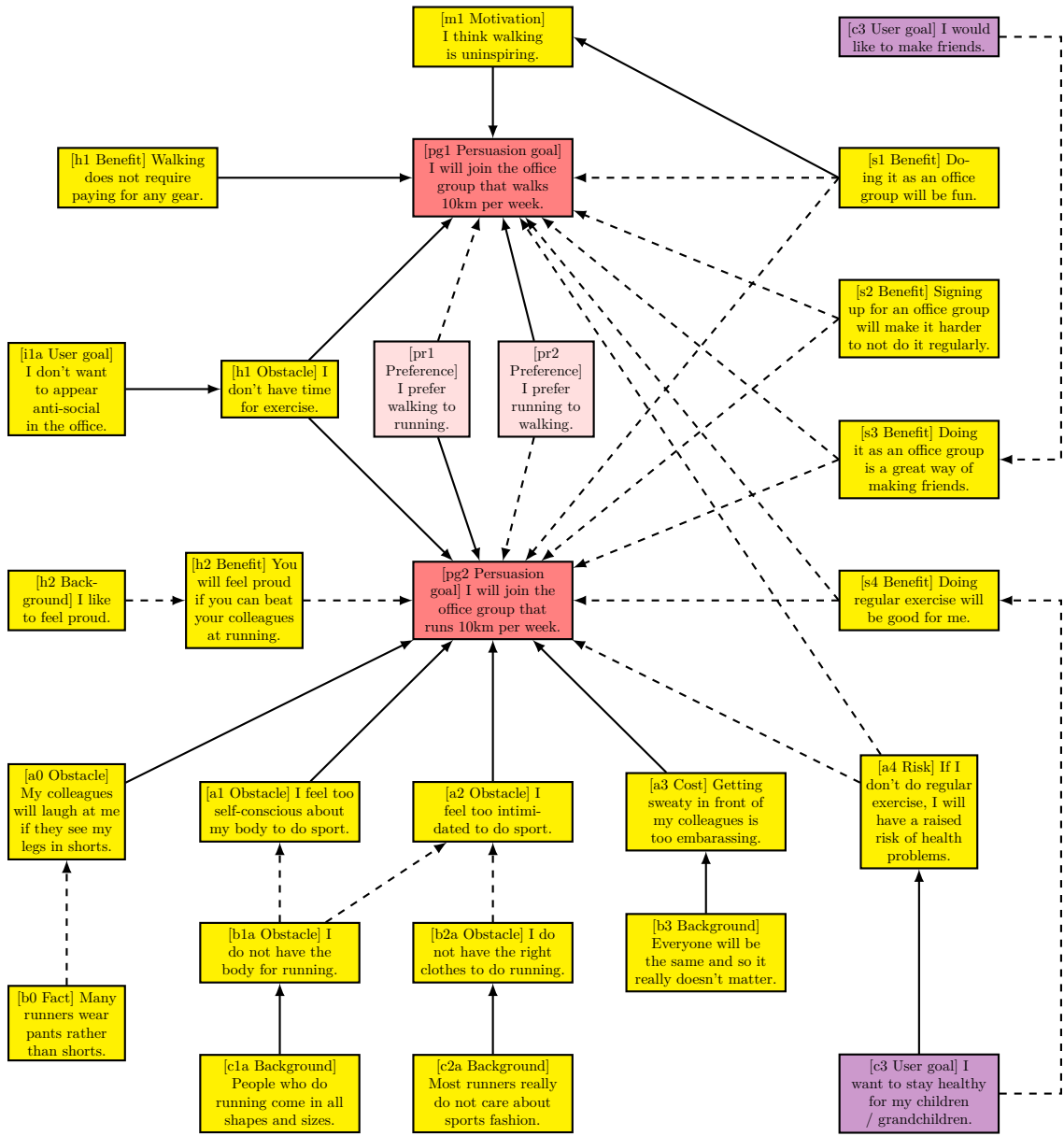

In developing the graphs in each case study, the ontological types were important in seeking the relevant arguments. As can be seen, there is a diverse range of ontological types appearing in the case studies.

Furthermore, each of the case studies has resulted in an argument graph with a mixture of functional types. For reasons of space, we only include one or two persuasion goals in each case study. In the three of the case studies, namely women in sport, office wellbeing initiative, and cervical screening, the large majority of arguments are subjective, whereas in the fourth case study, namely flu vaccination, there is an even balance of objective and subjective arguments. This is perhaps due to the availability of scientific evidence surrounding the effects of flu vaccination programmes.


Having the subtypes of the functional type helps in identifying the arguments and counterarguments. Each subtype can be treated as a question when searching for supporting or attacking arguments to any given argument. For instance, given the persuasion goal, what are the opinions that support or attack it, what are the objective arguments that support or attack it, what are the emotional arguments that support or attack it, what are the self-perception arguments that support or attack it, etc. Then for each argument that is put into the graph, this process can be repeated by recursion.

To illustrate the use of topic labelling, we can consider the argument graph for office wellbeing initiative given
in Figure \ref{fig:officegraph}. The key words are given in Figure \ref{fig:key} and the topic labellings are given in Figure \ref{fig:topics}. If required, a finer-grained topic hierarchy could be used to have more distinction between the arguments.

Finally, there are some context types for the case study arguments. In the women in sports case study, some arguments only apply to women (e.g., ``There are many women competing at the highest level in many different sports''), and some arguments only apply in countries where there is a government initiative for women in sports (e.g., ``With the government initiative, there are many opportunities to participate in women's sport in your area''). In the office wellbeing initiative, some arguments only apply to people who work (e.g., ``I would be too embarrassed to get sweaty in front of my colleagues''). In the cervical screening case study, most arguments only apply to women. In the flu vaccine for healthcare workers case study, the arguments that refer to staff only apply to NHS staff.


\section{Using domain models}
 \label{section:use}

The domain model can be used by the system to make strategic choices of move.
In order to illustrate this, we consider the office well-being initiative in Figure \ref{fig:officegraph}.
There are two persuasion goals here.

\begin{itemize}

\item (pg1) I will join the office group that walks 10km per week.

\item (pg2) I will join the office group that runs 10km per week.

\end{itemize}

Recall that the system has a user model. This is a model that the system has of the user. It may have gained information about the user by previous interactions with the user, or by asking the user questions, or by knowledge of agents similar to the user. We are not concerned here with how that model is obtained, we just make some assumptions about the information in it. In particular, for this example, we assume that the user model is a probability value for each argument. It is an estimate that the system has of the user's belief in each argument. Now, suppose that user model has a high probability of belief in the following argument.

\begin{itemize}

\item (pr2) I prefer running to walking.

\end{itemize}

Then according to the domain model, pg1 is attacked, and pg2 is supported. This suggests to the system, that strategically it would be better to try to persuade the user of pg2 (i.e., to join the office group that runs 10km per week).

Next, the system has some choices. Should it provide one or more opinions to support the persuasion goal (i.e one or more of s1 to s4), or should it provide an emotional argument (i.e., h2), or should it try to tackle the possible counterarguments that the user might have (i.e., a0 to a3), or should it try to support the persuasion goal with highlighting a future risk from not changing behaviour (i.e., a4).

To address this choice, the system again refers to the user model, and identifies that the user has a goal of making new friends (i.e., the user model has high belief in the argument c5). This supports s3, and so supports pg2. The system reminds the user of c5, and presents s3, as support for pg2.

\begin{itemize}

\item (c5) I would like to make friends.

\item (s3) Doing it [i.e., joining the group that runs 10km per week] as an office group is a great way of making friends.

\end{itemize}

When referring to the user model, the system may also notice that the user has some belief in a couple of counterarguments to the persuasion goal. These counterarguments are:

\begin{itemize}

\item (a0) My colleagues will laugh at me if they see my legs in shorts.

\item (a3) I would be too embarrassed to get sweaty in front of my colleagues.

\end{itemize}

Suppose the confidence for these beliefs is not high. According to the user model, the user believes these arguments, but the system is not confident that the user does indeed belief them. In other words, the system thinks the user model might be inaccurate with regard to these arguments for this user. To address this, the system can ask the user whether these arguments are reasons for the user to not accept the persuasion goal. Now, suppose the user says that a3 is indeed a reason to not accept the persuasion goal, the system can provide the following counterargument.

\begin{itemize}

\item (b3) Everyone will be the same and so it really doesn't matter.

\end{itemize}

This example illustrates how the domain model, together with the user model, can be used so that arguments are selected from the domain model as part of strategy to persuade the user.

As we have already mentioned, the tags can be further used to select certain kinds of argument in preference to others. For example, if in the user model it is known that the user responds better to supporting arguments than attacking arguments, or the user responds better to emotional arguments than factual arguments, then that strategic knowledge can be harnessed by the system when it is choosing which move to make.

\section{Discussion}
 \label{section:discussion}

Our primary concern in the paper is in domain modelling. For argument-based persuasion, we need access to relevant arguments and counterarguments. For this, we have made the following contributions:

\begin{itemize}

\item We have discussed how the beliefs of an agent concerning a healthcare problem is of central importance in judging whether a particular healthcare intervention is likely to be successful, and we have identified a number of dimensions for these beliefs that are considered implicitly or explicitly in the behaviour change literature (Section \ref{section:background}).

\item We have provided a framework for domain modelling that provides a number of types and subtypes. This framework forms the basis of questions that can be used to identify arguments and counterarguments for a specific behaviour change application, and it provides tags that can used by the computational persuasion system for optimizing the choice of argument or counterargument in a dialogue.

\item We evaluated the framework by constructing argument graphs for behaviour change applications in participation of women in sport, in office wellbeing initiative, in participation in cervical smear screening, and in flu vaccine for healthcare visitors. These case studies show how a wide range of the types of argument can appear each argument graph.

\end{itemize}

There are many approaches to domain modelling in computer science. However, to the best of our knowledge, there is no approach to domain modelling that is appropriate for acquiring and tagging different kinds of argument for use in behaviour change. Some of the types and subtypes do arise in agent-based modelling. In particular, the belief-desires-intentions (BDI) framework does provide an approach to modelling the differences between beliefs (i.e., what an agent believes about the world), desires (i.e., what an agent would like to achieve in the world), and intentions (i.e., what an agent has chosen to do in the world) \cite{Rao91}. Desires can lead to goals, and intentions can lead to plans. However, this framework is formalized in a modal logic that provides an axiomatization for representing and reasoning with these concepts.

As we suggested in Section \ref{section:computationalpersuasion}, the domain model is an important component of an automated persuasion system as it provides the all-important arguments for the dialogue. In addition, as we have suggested in Section \ref{section:use}, the domain model is important for making strategic choices of argument. Whilst the domain model is potentially valuable for diverse approaches to argumentation, we are particularly interested in using these models with the
epistemic approach to argumentation. This approach to probabilistic argumentation offers a formal framework for clarifying the nature of belief in arguments. It is based on a probability distribution over the subsets of arguments from which the belief in an argument is derived
\cite{Thimm12,Hunter2013ijar,BaroniGiacominVicig14,HunterThimm2016,PolbergHunterThimm2017}.
An argument can therefore be to a certain degree believed, disbelieved, or neither.
We have also undertaken empirical studies with participants that lend support to the value of probabilistic argumentation \cite{PolbergHunter2017}. 

We have applied the probabilistic approach to modelling a persuadee's beliefs in arguments. This includes methods 
for updating beliefs during a dialogue \cite{Hunter2015ijcai,Hunter2016sum}, 
for efficient representation and reasoning with the probabilistic user model \cite{HadouxHunter2016},
for representing uncertainty in the user model \cite{Hunter2016ecai},
for harnessing decision-theoretic decision rules for optimizing the choice of arguments based on the user model \cite{HadouxHunter2017}, 
and for crowd-sourcing belief in arguments that can be used to train classifiers to predict an agent's belief in arguments \cite{HunterPolberg2017}. 
These developments offer a framework with a well-understood theoretical methodology, and implementations that are computationally viable for strategic argumentation.

Alternative approaches to strategic argumentation include considering uncertainty with respect to the structure of the graph (i.e., the arguments the user is aware of) rather than the belief in arguments \cite{RTO13}, and with respect to previous moves the user has made \cite{HSMBM13}.

There are also some other investigations of computational models of argument with participants that lend support for probabilistic approach to argumentation, and for harnessing user models, in strategic argumentation. Studies performed by Rahwan {\em et al} \cite{Rahwan2011} and Cerruti {\em et al} \cite{Cerutti2014} investigated various forms of reinstatement in argumentation. The users were presented several argument graphs and were asked to explain how acceptable a given argument is in their opinion. The results show that in some cases, the implicit knowledge about domains can substantially affect the given acceptability levels. However, more importantly, the experiments show that the attacked argument's acceptability is lowered, but does not fall to $0$, which is what would be predicted by the usual dialectical semantics for abstract argumentation. Additionally, introducing the defense for this argument raises its acceptability. However, typically it does not reach the value of $1$, which is the level the usual dialectical semantics would predict. This study lends support for using the epistemic approach for a finer grained representation of belief/disbelief.

In a study of argumentation dialogues, Rosenfeld and Kraus \cite{Rosenfeld2015} undertook an experiment in order to develop a machine learning-based approach to predict the next move a participant would make in a dialogue. This work was further extended in \cite{RosenfeldKraus2016tis,RosenfeldKraus2016ecai}.
The machine learning models were trained on data that incorporated the sequences of arguments in a dialogue that the participants accept. Once trained, the models were able to predict the acceptance an unseen case would have.


There are also studies with participants by Masthoff and co-workers that investigate the efficacy of using arguments as a way of persuading people when compared with other counselling methods indicating that argumentation may have disadvantages if used inappropriately \cite{Nguyen2008}, and that rather than a confrontational approach, argumentation that is based on appeal to friends, appeal to group, or appeal to fun, may be more efficacious \cite{Vargheese2013,Vargheese2016}.




We now turn to the two inter-related topics that we would like to investigate in future work. These are the nature and role of emotion in argumentation, and the way that arguments are presented in argumentation.

Emotional arguments play on the emotions of the persuadee. They attempt to invoke an emotional response in the user in way that she would be persuaded by the argument. An emotional argument is often an argument that involves accepting, or believing, or doing, something will cause a positive or negative emotion in you or those around you.

\begin{itemize}
\item You will be elated if you complete the marathon.
\item Taking the flu vaccine means that you do not need to worry about getting flu this year.
\item Many of your colleagues are doing exercise classes, and so you will feel left out if you don't participate.
\end{itemize}

Emotional arguments are predominantly deployed by the system in order to influence the user via emotional devices to be persuaded to agree to the persuasion goal. In certain situations, they can be powerful arguments in persuasion. For instance, Lukin {\em et al} \cite{Lukin2017} have shown that with some audiences, emotional arguments are more effective in persuasion than factual arguments. For this, they categorized audiences according to the OCEAN personality traits (i.e., openness to experience, extroversion, agreeableness, conscientiousness, and neuroticism --- see \cite{Goldberg93} for a review of personality traits).

Emotion in argumentation has also been the subject of a study with participants in a debate where the emotional state was estimated from EEG data and automated facial expression analysis. In this study, Benlamine {\em et al} \cite{Benlamine2015} showed for instance that the number and the strength of arguments, attacks and supports exchanged between a participant could be correlated with particular emotions of the participant.

However, the modelling emotional aspects of argument has received little attention in the computational argumentation literature. There is a proposal for rules for specifying scenarios where empathy is given or received in negotiation \cite{Martinovski2009}, and there is a proposal for specifying argument schema (rules that specify general patterns of reasoning) for capturing aspects of emotional argument \cite{LloydKelly2012}. It would be interesting to investigate if how these ideas could be adapted to the context of persuasion dialogues, and in particular how this might impinge upon the tagging of the emotional dimensions as suggested in this paper.


There are various ontologies for emotion types. For instance, the emotion annotation and representation language (EARL) proposed by the Human-Machine Interaction Network on Emotion (HUMAINE) classifies 48 emotions \cite{Schroder2006,Schroder2007}.
In general, the structure of such ontologies is debatable. Nonetheless, they may be useful in labelling arguments with respect to the kind of emotion is being caused by the statement of the argument. To illustrate this, we consider the following arguments taken from our case study:

\begin{itemize}
\item You will feel guilty if you do not do the [smear] test.
\item Think how you would feel if you only found out the you had it [i.e., cervical cancer] once it had advanced.
\end{itemize}

In the first argument, the decision to not do the smear test is explicitly claimed to cause a feeling of guilt (a negative thought according to the EARL ontology), whereas in the second argument, the invoking of an emotional response is implicit. In the second argument, the words ``think how you would feel'' together with the negative outcome is meant to invoke a negative feeling in the reader.

Whilst it may be possible to automatically identify emotional arguments using sentiment analysis technology, the variety and subtlety of emotion arguments means that in the short-term annotators would be required to identify emotional arguments. It would appear feasible to identify a range of general patterns of emotional arguments that could be incorporated in guidelines for annotation. It is interesting to note that affective computing has put emotion at the centre of the relationship between users and computing systems \cite{Calvo2010}. It therefore seems that affective computing offers a number of ways to build on the suggestion for tagging of emotional arguments as suggested here.

The way that arguments are presented to users can affect the success with which they can persuade the user. This may depend on the emotional response invoked, and this may in turn be related to the personality of the user.
Personality is clearly an important aspect of persuasion \cite{Nahai2017}.
Some arguments only work with certain kinds of people --- for instance, for an argument to persuade someone to vote in the national election we could have the following arguments:

\begin{itemize}
\item If the person ``follows the crowd'',
then use an argument telling her that the majority of the population voted in the last election.
\item If the person ``follows rules rigorously'',
then use an argument telling her that it is her duty to vote.
\end{itemize}

Above suggests that arguments could be classified/tagged by the types of personality that might respond positively to them. For example, arguments could be tagged according to what kind of appeal they have (e.g., to authority, to conform to rules, to follow the crowd, to win approval, etc.). Then, when the system is choosing which arguments to use, it could select arguments that are more likely to be efficacious with the personality of that particular user. To classify the user, standard methods for personalty traits could be used (e.g., OCEAN \cite{Goldberg93}) or multi-dimensional private traits could be predicted from digital records of the user (e.g., from Facebook likes) \cite{Kosinski2013}. Then different dimensions for typing arguments according to presentation style include the following.






\begin{itemize}

\item The {\bf grammatical style} concerns the grammatical phrasing of the premises and claim of the argument. This includes whether the verb is imperative v. conditional, whether first v. third person is used, etc.

\begin{itemize}
\item You should take more exercise.
\item You are advised to take more exercise.
\item It would be good if you took more exercise.
\item If I were you, I would take more exercise.
\item Everyone is recommended to consider taking more exercise.
\end{itemize}

Different kinds of user may respond differently to different phraseology. For instance, an imperative style may work with some kinds of user, and be counter-productive with others.

\item The {\bf linguistic style} concerns the broader linguistic style of the premises and claim of the argument. For instance, whether an informal or formal style is used. For this, the choice of words, including colloquial words, the complexity of sentences, etc., can be important. As an example, emotion invoked through choice of words can be important in persuasion. This is a well-studied area in psychology through the development of effective norm databases \cite{Bradley1999,Warriner2013}. These capture the emotion response to specific words in three dimensions: arousal (ranging from excited to calm), valence (ranging from pleasant to unpleasant), and dominance (ranging from being in control to being dominated). To illustrate, for valence scores, {\em leukemia} and {\em murder} are low and {\em sunshine} and {\em lovable} are high; for arousal scores, {\em grain} and {\em dull} are low and {\em lover} and {\em terrorism} are high; and for dominance scores, {\em dementia} and {\em earthquake} are low, and {\em smile} and {\em completion} are high.


\item The {\bf framing style} of an argument concerns the exact phrasing of an argument which can influence how it is perceived. It is increasingly acknowledged in healthcare professions that interactions with patients should use effective communication that is respectful and non-judgmental (see, for example, \cite{NICE2014}). Such a stance would be a natural point for automated persuasion systems. However, this still leaves much latitude in how arguments can be framed. Consider, for example, experiments by Tversky and Kahneman where people face one of the two situations \cite{Tversky1981science}:

\begin{itemize}

\item Situation 1: Vaccine A can only save 200 lives for sure, whereas vaccine B has a 1/3 probability of saving 600 lives and a 2/3 probability of saving none.

\item Situation 2: Vaccine C causes 400 people to die for sure, whereas with vaccine D there is a 1/3 probability that nobody dies, and a 2/3 probability that 600 people die.

\end{itemize}

Obviously, the argument is the same in both situations if we consider a population of 600.
However, in situation 1, people massively prefer vaccine A while in situation 2 they prefer vaccine D.
The explanation is that in the context of gain (situation 1) people tend to be averse to risk and in the context of loss (situation 2) they try to minimize the loss.

\end{itemize}

In a user study on the persuasiveness of healthy eating messages \cite{JosekuttyThomas2017}, positively framed messages (e.g., Most people believe that eating a healthy breakfast contributes to a longer lifespan) were considered better than negatively framed messages (e.g., Most people believe that eating an unhealthy breakfast contributes to a shorter lifespan). Furthermore, Cialdini's principles of persuasion \cite{Cialdini84} were considered (i.e., reciprocation, commitment, consensus, liking, authority, and scarcity), and it was found that arguments that appeal to authority (e.g., Studies conducted by health experts have shown that eating a healthy breakfast keeps you energized) are the most persuasive.

The need to recognize the primary audience of healthcare messages is well understood in healthcare communications in general \cite{OSullivan03}, that the messages are chosen to be easy to process \cite{Suggs15}, and ideally that are individualized messages \cite{Dimarco06}. The above suggestions for future work go beyond this by developing techniques for argumentation that type the arguments according to the types of emotion likely to be invoked by the argument, and the types of user according to the personality for whom the argument will be effective. Since, the arguments are presented in natural language, this then raises the need to better understand how different presentational styles could be incorporated in the domain modelling.


\subsubsection*{Acknowledgments}

This research is funded by EPSRC grant EP/N008294/1 {\em Framework for Computational Persuasion}.


\newcommand{\etalchar}[1]{$^{#1}$}

\end{document}